\documentclass[twoside]{article}
\usepackage[accepted]{aistats2022}

\usepackage[round]{natbib}

\usepackage{parskip}
\usepackage[utf8]{inputenc} 
\usepackage[T1]{fontenc}    
\usepackage{hyperref}       
\usepackage{url}            
\usepackage{booktabs}       
\usepackage{amsfonts}       
\usepackage{nicefrac}       
\usepackage{microtype}      
\usepackage{xcolor}         

\usepackage{algorithm}
\usepackage[noend]{algpseudocode}
\usepackage{tikz,lipsum,lmodern}
\usepackage[most]{tcolorbox}
\usepackage{xspace}
\usepackage{bm}
\usepackage{amssymb}

\newcommand{\data}{\mathcal{D}}
\newcommand{\post}{p\left(\theta|\data\right)}
\newcommand{\samples}{\mathcal{S}}
\newcommand{\subsamples}{\mathcal{S}^\prime}
\newcommand{\subsamplestwo}{\mathcal{S}^{\prime\prime}}
\newcommand{\subsamplesthree}{\mathcal{S}^{\prime\prime\prime}}
\renewcommand{\lq}{\mathcal{L}}
\newcommand{\ninit}{n_0}
\newcommand{\X}{\mathcal{X}}

\newcommand{\appropto}{\mathrel{\vcenter{
  \offinterlineskip\halign{\hfil$##$\cr
    \propto\cr\noalign{\kern2pt}\sim\cr\noalign{\kern-2pt}}}}}

\algnewcommand\algorithmicinput{\textbf{Input:}}
\algnewcommand\INPUT{\item[\algorithmicinput]}
\algloop{Do}

\algblock{ParFor}{EndParFor}
\algnewcommand\algorithmicparfor{\textbf{parfor}}
\algnewcommand\algorithmicpardo{\textbf{do}}
\algnewcommand\algorithmicendparfor{\textbf{end\ parfor}}
\algrenewtext{ParFor}[1]{\algorithmicparfor\ #1\ \algorithmicpardo}
\algrenewtext{EndParFor}{\algorithmicendparfor}
\usepackage{xspace}
\usepackage{ifthen}
\usepackage{todonotes}

\usepackage{hyperref}
\hypersetup{
    linkbordercolor=cyan,
    urlbordercolor=magenta,
    citebordercolor=green,
}

\newcommand{\meanstd}[3][]{
    \ifthenelse{ \equal{#1}{} }
    {#2 {\scriptstyle \pm #3}}
    {#2 {\scriptstyle \pm #3}\times10^{#1}}
}
\newcommand{\gskl}{GsKL\xspace}

\newcommand{\nparams}{D}
\newcommand{\normpdf}[3]{\mathcal{N}\left({#1}; {#2}, {#3 } \right)}
\newcommand{\x}{x}
\newcommand{\xx}{\mathbf{X}}
\newcommand{\y}{\mathbf{y}}
\newcommand{\vgp}{\psi}
\newcommand{\covar}{C}
\newcommand{\sigmalik}{\sigma}

\begin{document}

\twocolumn[

\aistatstitle{Parallel MCMC Without Embarrassing Failures}

\aistatsauthor{Daniel Augusto de Souza$^1$, Diego Mesquita$^{2,3}$, Samuel Kaski$^{2,4}$, Luigi Acerbi$^5$}

\aistatsaddress{ $^1$University College London  $^2$Aalto University $^3$Getulio Vargas Foundation\\ $^4$University of Manchester $^5$University of Helsinki\\ \texttt{\small daniel.souza.21@ucl.ac.uk, diego.mesquita@fgv.br, samuel.kaski@aalto.fi,  luigi.acerbi@helsinki.fi} }]

\runningauthor{Daniel Augusto de Souza, Diego Mesquita, Samuel Kaski, Luigi Acerbi}

\begin{abstract}
\emph{Embarrassingly parallel} Markov Chain Monte Carlo (MCMC) exploits parallel computing to scale Bayesian inference to large datasets by using a two-step approach. First, MCMC is run in parallel on (sub)posteriors defined on data partitions. Then, a server combines local results. While efficient, this framework is very sensitive to the quality of subposterior sampling. Common sampling problems such as missing modes or misrepresentation of low-density regions are amplified  -- instead of being corrected -- in the combination phase, leading to catastrophic failures.
In this work, we propose a novel combination strategy to mitigate this issue. Our strategy, Parallel Active Inference (PAI), leverages Gaussian Process (GP) surrogate modeling and active learning. After fitting GPs to subposteriors, PAI (i) shares information between GP surrogates to cover missing modes; and (ii) uses active sampling to individually refine subposterior approximations.  We validate PAI in challenging benchmarks, including heavy-tailed and multi-modal posteriors and a real-world application to computational neuroscience. Empirical results show that PAI succeeds where previous methods catastrophically fail, with a small communication overhead.
\end{abstract}

\section{INTRODUCTION}

Markov Chain Monte Carlo (MCMC) methods have become a gold standard in Bayesian statistics \citep{gelman2013bayesian,Carpenter2017}.
However, scaling MCMC methods to large datasets is challenging due to their sequential nature and that they typically require many likelihood evaluations, implying repeated sweeps through the data.
Various approaches that leverage distributed computing have been proposed to mitigate these limitations \citep{Angelino+others:2016, Robert2018}.
In general, we can split these approaches between those that incur constant communication costs and those requiring frequent interaction between server and computing nodes \citep{vehtari2020expectation}.

\emph{Embarrassingly parallel} MCMC~\citep{Neiswanger2014} is a popular class of methods which employs a divide-and-conquer strategy to sample from a target posterior, requiring only a single communication step. For dataset $\mathcal{D}$ and model parameters $\theta \in \mathbb{R}^D$, suppose we are interested in the  Bayesian posterior $\post \propto p(\theta) p(\mathcal{D} | \theta)$, where $p(\theta)$ is the prior and $p(\mathcal{D} | \theta)$ the likelihood. Embarrassingly parallel methods begin by splitting the data $\mathcal{D}$ into $K$ smaller partitions $\mathcal{D}_1, \ldots, \mathcal{D}_K$ so that we can rewrite the posterior as
\vspace{-0.25em}
\begin{equation}\label{eq:pde}
    \post \propto \prod_{k=1}^{K} p(\theta)^{1/K} p(\data_k | \theta) \equiv
    \prod_{k=1}^{K} p_k(\theta).
\vspace{-0.2em}
\end{equation}
Next, an MCMC sampler is used to draw samples $\samples_k$ from each \emph{subposterior} $p_k(\theta)$, for $k=1\ldots K$, in parallel. Then, the computing nodes send the local results to a central server, for a final aggregation step. These local results are either the samples themselves or approximations $q_1, \ldots, q_K$ built using them.

Works in embarrassingly parallel MCMC mostly focus on combination strategies.
\citet{Scott2016} employ a weighted average of subposterior samples.
\citet{Neiswanger2014} propose using multivariate-normal surrogates  as well as non-parametric and semi-parametric forms.
\citet{Wang2015} combine subposterior samples into a hyper-histogram with random partition trees.
\citet{Nemeth2018} leverage density values computed during MCMC to fit Gaussian process (GP) surrogates to log-subposteriors.
\citet{Mesquita2019} use subposterior samples to fit normalizing flows and apply importance sampling to draw from their product.

Despite these advances, parallel MCMC suffers from an unaddressed limitation: its dependence on high-quality subposterior sampling.
This requirement is especially difficult to meet when subposteriors are multi-modal or heavy-tailed, in which cases MCMC chains often visit only a subset of modes and may underrepresent low-density regions.
Furthermore, the surrogates $(q_k)_{k=1}^{K}$ built only on local MCMC samples might match poorly the true subposteriors if not carefully tuned.
%
%

\paragraph{Outline and contributions.}  We first discuss the failure modes of parallel MCMC (Section \ref{sec:embarrassing}).
Drawing insight from this discussion, Section \ref{sec:method} proposes a novel GP-based solution, Parallel Active Inference (PAI).
After fitting the subposterior surrogates, PAI shares a subset of samples between computing nodes to prevent mode collapse.
PAI also uses active learning to refine low-density regions and avoid catastrophic model mismatch.
Section \ref{sec:experiments} validates our method on challenging benchmarks and a real-world neuroscience example.
Finally, Section \ref{sec:related_works} reviews related work and Section \ref{sec:discussion} discusses strengths and limitations of PAI.

\section{EMBARRASSINGLY PARALLEL MCMC: HOW CAN IT FAIL?}
\label{sec:embarrassing}

\begin{figure*}[tb!]
\centering
\includegraphics[width=0.9\linewidth]{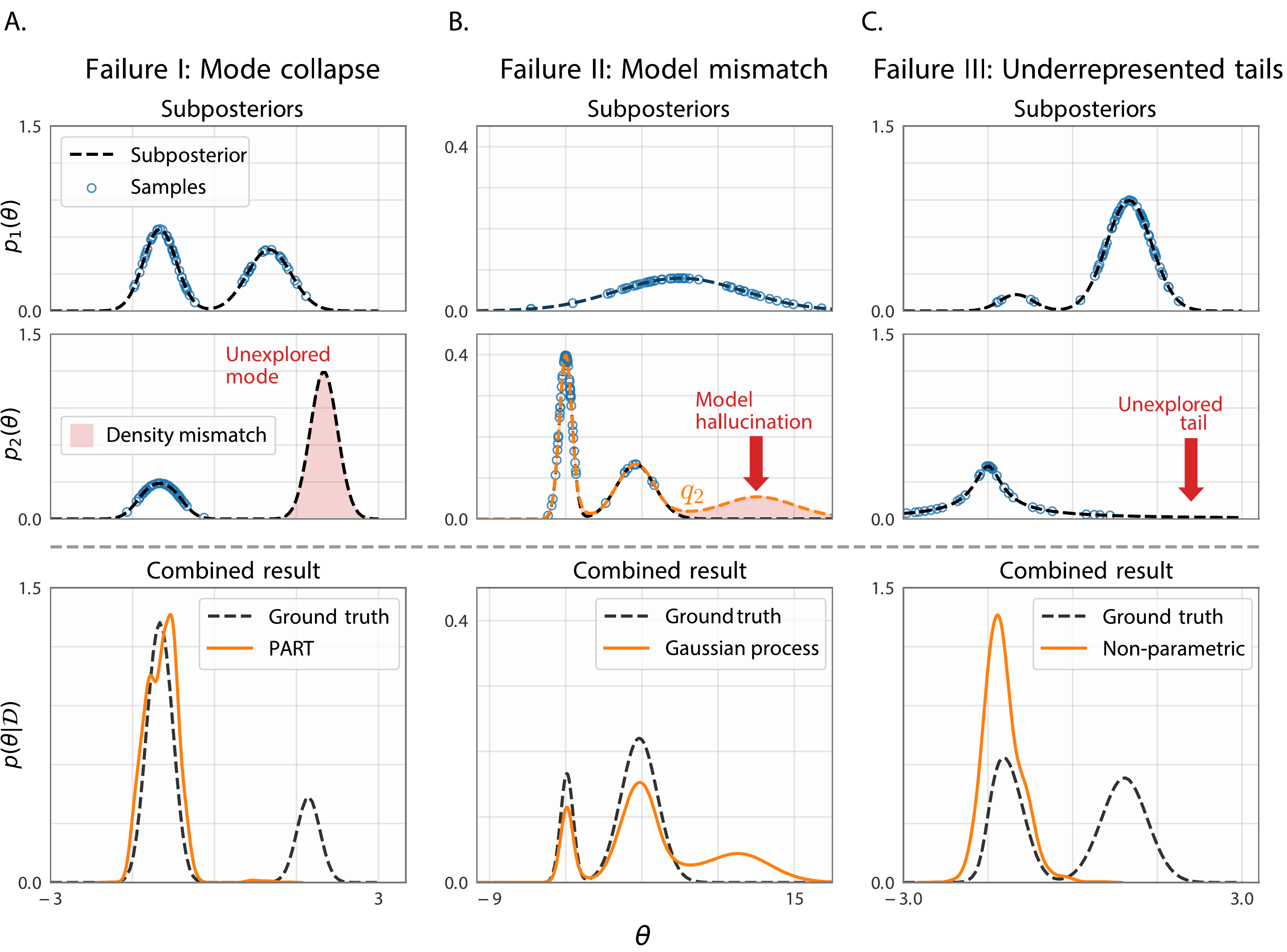}
  \caption{{\bf Failure modes of embarrassingly parallel MCMC.} \textbf{A--C}. Each column illustrates a distinct failure type described in Section \ref{sec:failures}. For each column, the top rows show two subposteriors $p_k(\theta)$ (black dashed line: ground truth; blue circles: MCMC samples), and the bottom row shows the full posterior $p(\theta|\mathcal{D})$ with the approximate posterior combined using the method specified in the legend (orange line). These failure modes are general and not unique to the displayed algorithms (see Appendix \ref{supp:failure_modes} %
  for details and further explanations).
  }
  \label{fig:toy}
\end{figure*}


We recall the basic structure of a generic embarrassingly parallel MCMC algorithm in Algorithm \ref{alg:basic}. 
%
%
This schema has major failure modes that we list below, before discussing potential solutions. We also illustrate these pathologies in Fig \ref{fig:toy}.
In this paper, for a function $f$ with scalar output and a set of points $\samples = \{s_1, \ldots, s_N \}$, we denote with $f(\samples) \equiv \{f(s_1), \ldots, f(s_N)\}$.

\begin{algorithm}[h!]
\caption{Generic embarrassingly parallel MCMC}\label{alg:basic}
\begin{algorithmic}[1] \small
\INPUT Data partitions $\data_1, \ldots, \data_K$; prior $p(\theta)$; likelihood function $p(\data|\theta)$. 
\ParFor{$1 \ldots K$} \Comment{Parallel steps}
\State $\samples_k \leftarrow $ MCMC samples from $p_k(\theta) \propto p(\theta)^{1/K} p(\data_k | \theta)$
\State build subposterior model $q_k(\theta)$ from $\samples_k$ 
\EndParFor
\State Combine: $q(\theta) \propto \prod_{k=1}^K q_k(\theta) $ \Comment{Centralized step}
\end{algorithmic}
\end{algorithm}

\subsection{Failure modes}
\label{sec:failures}

\paragraph{I: Mode collapse.}

It is sufficient that \emph{one} subposterior $q_k$ misses a mode for the combined posterior $q$ to lack that mode (see Fig \ref{fig:toy}A). While dealing with multiple modes is an open problem for MCMC, here the issue is exacerbated by the fact that a single failure propagates to the final solution. A back-of-the-envelope calculation shows that even if the chance of missing a mode is small, $\varepsilon > 0$, the probability of mode collapse in the combined posterior is $\approx (1 - \varepsilon)^K$ making it a likely occurrence for sufficiently large $K$.

\begin{tcolorbox}[colback=red!5!white,colframe=red!75!black]
\textbf{Insight 1:}
For multimodal posteriors, mode collapse is almost inevitable unless the computing nodes can exchange information about the location of important posterior regions.
\end{tcolorbox}

\paragraph{II: Catastrophic model mismatch.}

Since the $q_k$ are approximations of the true subposteriors $p_k$, small deviations between them are expected -- this is not what we refer to here. Instead, an example of \emph{catastrophic} model mismatch is when a simple parametric model such as a multivariate normal is used to model a multimodal posterior with separate modes (see Section \ref{sec:experiments}).
Even nonparametric methods can be victims of this failure.
For example, GP surrogates are often used to model nonparametric deviations of the log posterior from a parametric `mean function'. While these models can well represent multimodal posteriors, care is needed to avoid grossly mismatched solutions in which a $q_k$ `hallucinates' posterior mass due to an improper placement of the GP mean function (Fig \ref{fig:toy}B).
%

\begin{tcolorbox}[colback=red!5!white,colframe=red!75!black]
\textbf{Insight 2:} We cannot take subposterior models at face value. Reliable algorithms should check and refine the $q_k$'s to avoid catastrophic failures.
\end{tcolorbox}

\paragraph{III: Underrepresented tails.} This effect is more subtle than the failure modes listed above, but it contributes to accumulating errors in the estimate of the combined posterior. The main issue here is that, by construction, MCMC samples and subposterior models based on these samples focus on providing information about the high-posterior-mass region of the subposterior. However, different subposteriors may overlap only in their tail regions (Fig \ref{fig:toy}C), implying that the tails and the nearby `deep' regions of each subposterior might actually be the most important in determining the exact shape of the combined posterior.

\begin{tcolorbox}[colback=red!5!white,colframe=red!75!black]
 \textbf{Insight 3:} Subposterior models built only from MCMC samples (and their log density) can miss important information about the tails and nearby regions of the subposterior which would also contribute to the combined posterior.
\end{tcolorbox}


\subsection{Past solutions}
\label{sec:past}

Since there is no guarantee that $q$ approximates well the posterior $p$, \citet{Nemeth2018} refine $q$ with an additional parallel step, called Distributed Importance Sampler (DIS). DIS uses $q$ as a proposal distribution and draws samples $\samples \sim q$, that are then sent back for evaluation of the log density $\log p_k(\samples)$ at each parallel node. The true log density $\log p(\samples) = \sum_k \log p_k(\samples)$ is then used as a target for \emph{importance sampling/resampling}  \citep{robert2013monte}. Technically, this step makes the algorithm not `embarrassingly parallel' anymore, but the prospect of fixing catastrophic failures outweighs the additional communication cost. However, DIS does not fully solve the issues raised in Section \ref{sec:failures}. Notably, importance sampling will not help recover the missing regions of the posterior if $q$ does not cover them in the first place. DIS can help in some model mismatch cases, in that `hallucinated' regions of the posterior will receive near-zero weights after the true density is retrieved.

\subsection{Proposed solution}
\label{sec:solutions}

\begin{figure*}[t!]
\centering
\includegraphics[width=0.95\linewidth]{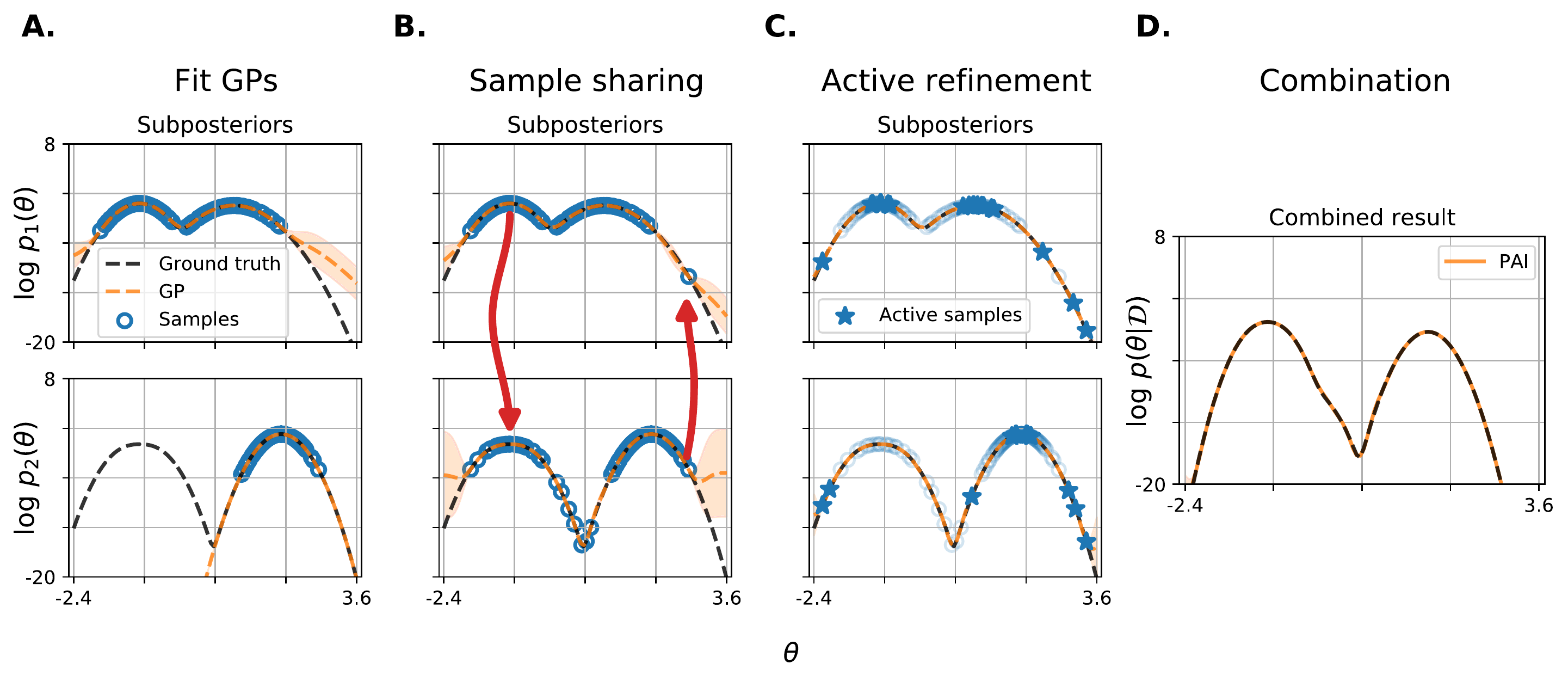}
\vspace{-1em}
  \caption{{\bf Parallel active inference (PAI).} \textbf{A.} Each log subposterior $\log p_k(\theta)$ (black dashed line) is modeled via Gaussian process surrogates (orange dashed line: mean GP; shaded area: 95\% confidence interval) trained on MCMC samples $\mathcal{S}_k$ (blue circles) and their log-density $\log p_k(\mathcal{S}_k)$. Here, MCMC sampling on the second subposterior has missed a mode. \textbf{B.} Selected subsets of MCMC samples are shared across nodes, evaluated locally and added to the GP surrogates. Here, sample sharing helps finding the missing mode in the second subposterior, but the GP surrogate is now highly uncertain outside the samples. \textbf{C.} Subposteriors are refined by actively selecting new samples (stars) that resolve uncertainty in the surrogates. \textbf{D.} Subposteriors are combined into the full approximate log posterior (orange line); here a perfect match to the true log posterior (black dashed line).
  \label{fig:sols}
  }
\end{figure*}

Drawing from the insights in Section \ref{sec:failures}, we propose two key ideas to address the blind spots of embarrassingly parallel MCMC. Here we provide an overview of our solution, which is described in detail in Section \ref{sec:method}. The starting point is modeling subposteriors via Gaussian process surrogates (Fig \ref{fig:sols}A).

\paragraph{Sample sharing.} We introduce an additional step in which each node shares a selected subset of MCMC samples with the others (Fig \ref{fig:sols}B). This step provides sufficient information for local nodes to address mode collapse and underrepresented tails. While this communication step makes our method not strictly `embarrassingly' parallel, we argue it is necessary to avoid posterior density collapse. Moreover, existing methods already consider an extra communication step \citep{Nemeth2018}, as mentioned in Section \ref{sec:past}.

\paragraph{Active learning.} We use \emph{active learning} as a general principle whenever applicable. The general idea is to select points that are informative about the shape of the subposterior, minimizing the additional communication required.
Active learning is used here in multiple steps: when selecting samples from MCMC to build the surrogate model (as opposed to thinning or random subsampling); as a way to choose which samples from other nodes to add to the current surrogate model of each subposterior $q_k$ (only informative samples are added); to actively sample \emph{new} points to reduce uncertainty in the local surrogate $q_k$ (Fig \ref{fig:sols}C). Active learning contributes to addressing both catastrophic model mismatch and underrepresented tails.

Combined, these ideas solve the failure modes    discussed previously (Fig \ref{fig:sols}D).

\section{PARALLEL ACTIVE INFERENCE}
\label{sec:method}

In this section, we present our framework, which we call Parallel Active Inference (PAI), designed to address the issues discussed in Section \ref{sec:embarrassing}. The steps of our method are schematically explained in Fig \ref{fig:sols} and the detailed algorithm is provided in Appendix \ref{supp:algorithm}.

\subsection{Subposterior modeling via GP regression}
\label{sec:gpsubposteriors}

As per standard embarrassingly parallel algorithms, we assume each node computes a set of samples $\samples_k$ and their log density, $\log p_k(\samples_k)$, by running MCMC on the subposterior $p_k$. 
We model each log-subposterior $\lq_k(\theta) \equiv \log q_k(\theta)$ using GP regression (Fig \ref{fig:sols}A; see \cite{rasmussen2006gaussian,Nemeth2018,gortler2019visual} and Appendix \ref{supp:gps} for more information).
We say that a GP surrogate model is trained on $\samples_k$ as a shorthand for $\left(\samples_k, \log p_k(\samples_k) \right)$.

When building the GP model of the subposterior, it is not advisable to use all samples $\samples_k$ because: (1) exact inference in GPs scales cubically in the number of training points (although see \cite{wang2019exact}); (2) we want to limit communication costs when sharing samples between nodes; (3) there is likely high redundancy in $\samples_k$ about the shape of the subposterior. \citet{Nemeth2018} simply choose a subset of samples by `thinning' a longer MCMC chain at regular intervals. Instead, we employ \emph{active subsampling} as follows.

First, we pick an initial subset of $\ninit$ samples $\samples_k^{(0)} \subset \samples_k$, that 
we use to train an initial GP (details in Appendix \ref{sec:step1}). Then, we iteratively select points $\theta^\star$ from $\samples_k$ by maximizing the \emph{maximum interquantile range} (MAXIQR) acquisition function \citep{jarvenpaa2021parallel}:
\begin{equation} \label{eq:maxiqr}
 \begin{split}
\theta^\star  
    &= \arg\max_\theta \left\{ e^{m(\theta; \samples_k^{(t)})} \text{sinh}  \left(u \cdot s(\theta; \samples_k^{(t)})\right)\right\},
\end{split}
\end{equation}
where $m(\theta; \samples_k^{(t)})$ and $s(\theta; \samples_k^{(t)})$ are, respectively, the posterior latent mean and posterior latent standard deviation of the GP at the end of iteration $t \geq 0$;
%
%
and $\sinh(z) = (\exp(z) -\exp(-z))/2$ for $z \in \mathbb{R}$ is the hyperbolic sine.
Eq. \ref{eq:maxiqr} promotes selection of points with high posterior density for which the GP surrogate is also highly uncertain, with the trade-off controlled by $u > 0$, where larger values of $u$ favor further exploration. In each iteration $t+1$, we greedily select a batch of $n_\text{batch}$ points at a time from $\samples_k \setminus \samples_k^{(t)}$ using a batch version of MAXIQR \citep{jarvenpaa2021parallel}. We add the selected  points to the current training set, $\samples_k^{(t+1)}$, and retrain the GP after each iteration (see Appendix \ref{sec:step1}). 
After $T$ iterations, our procedure yields a  subset of points $\subsamples_k \equiv \samples_k^{(T)} \subseteq \samples_k$ that are highly informative about the shape of the subposterior.

\subsection{Sample sharing}
\label{sec:samplesharing}

In this step, each node $k$ shares the selected samples $\subsamples_k$ with all other nodes (Fig \ref{fig:sols}B). Thus, node $k$ gains access to the samples $\subsamples_{\setminus k} \equiv \bigcup_{j \neq k} \subsamples_j$. Importantly, $\subsamples_{\setminus k}$ might contain samples from relevant
subposterior regions that node $k$ has has not explored.
As discussed in Section \ref{sec:gpsubposteriors}, for efficiency we avoid adding \emph{all} points $\subsamples_{\setminus k}$ to the current GP surrogate for subposterior $k$.
Instead, we add a sample $\theta^\star \in \subsamples_{\setminus k}$ to the GP training set only if the prediction of the current GP surrogate deviates from the true subposterior $\log p_k(\theta^\star)$ in a significant way (see Appendix \ref{sec:step2} for details).
%
After this step, we obtain an expanded set of points $\subsamplestwo_k$ that includes information from all the nodes, minimizing the risk of mode collapse (see Section \ref{sec:failures}). 

\subsection{Active subposterior refinement}
\label{sec:activerefinement}

So far, the GP models have been trained using selected subsets of samples from the original MCMC runs. In this step, we refine the GP model of each subposterior by sampling \emph{new} points (Fig \ref{fig:sols}C). Specifically, each node $k$ actively selects new points by optimizing the MAXIQR acquisition function (Eq. \ref{eq:maxiqr}) over $\mathcal{X} \subseteq \mathbb{R}^D$ (see Appendix \ref{sec:step3}). New points are selected greedily in batches of size $n_\text{batch}$, retraining the GP after each iteration.
This procedure yields a refined set of points $\subsamplesthree_k$ which includes new points that better pinpoint the shape of the subposterior, reducing the risk of catastrophic model mismatch and underrepresented tails. The final log-subposterior surrogate model $\lq_k$ is the GP trained on $\subsamplesthree_k$. 

\subsection{Combining the subposteriors}
\label{sec:combining}

Finally, we approximate the full posterior $\log \post = \sum_{k=1}^K \log p_k(\theta)$ by combining all subposteriors together (Fig \ref{fig:sols}D). Since each log-subposterior is approximated by a GP, the approximate full log-posterior is a sum of GPs and itself a GP, $\lq(\theta) = \sum_{k=1}^K \lq_k(\theta)$. Note that $\lq(\theta)$, being a GP, is still a distribution over functions. We want then to obtain a point estimate for the (unnormalized) posterior density corresponding to $\exp \lq(\theta)$. One choice is to take the posterior mean, which leads to the expectation of a log-normal density \citep{Nemeth2018}. We prefer a robust estimate and use the posterior median instead \citep{jarvenpaa2021parallel}. Thus, our estimate is
\begin{equation} \label{eq:sumgp}
    q(\theta) \propto \exp\left\{\sum_{k=1}^K m_k(\theta; \subsamplesthree_k) \right\}.
\end{equation}
In low dimension ($D = 1,2$), Eq. \ref{eq:sumgp} can be evaluated on a grid. In higher dimension,
one could sample from $q(\theta)$ using MCMC methods such as NUTS \citep{hoffman2014no}, as done by \cite{Nemeth2018}. However, $q(\theta)$ is potentially multimodal which does not lend itself to easy MCMC sampling.
%
Alternatively, \citet{acerbi2018variational} runs variational inference on $q(\theta)$ using as variational distribution a mixture of Gaussians with a large number of components. Finally, for moderate $D$, importance sampling/resampling with an appropriate (adaptive) proposal would also be feasible.

As a final optional step, after combining the subposteriors into the full approximate posterior $q(\theta)$, we can refine the solution using \emph{distributed importance sampling} (DIS) as proposed by \citet{Nemeth2018} and discussed in Section \ref{sec:past}.

\subsection{Complexity analysis}

Similarly to conventional embarrassingly parallel MCMC, we can split the cost of running PAI into two main components. The first consists of local costs, which involve computations happening at individual computing nodes (i.e., model fitting and active refinement). The second are global (or aggregation) costs, which comprise communication and sampling from the combined approximate posterior.

\subsubsection{Model fitting}

After sampling their subposterior, each computing node $k$ has to fit the surrogate model on the subset of their samples, $\subsamples_k$. These subsets are designed such that their size is $\mathcal{O}(D)$ (see Appendix \ref{supp:algorithm}). Thus, the cost of fitting the surrogate GP models in each of the $K$ computing nodes is $\mathcal{O}(D^3)$. The same applies for $\subsamplestwo_k$ and $\subsamplesthree_k$.

\subsubsection{Communication costs}
Traditional embarrassingly parallel MCMC methods only require two global communication steps: (1) the central server splits the $N$ observations among $K$ computing nodes; (2) each node sends $S$ subposterior samples of dimension $D$ back to the server, assuming nodes draw the same number of samples. Together, these steps amount to $\mathrm{O}(N + K S D)$ communication cost.

PAI imposes another communication step, in which nodes share their subposterior samples and incurring $\mathrm{O}(K S D)$ cost. Furthermore, supposing PAI acquires $A$ active samples to refine each subposterior, the cost of sending local results to servers is increased by $\mathrm{O}(KAD)$. PAI also incur a small additional cost for sending back the value of the GP hyperparameters $\mathrm{O}(KD)$. In sum, since usually $A \ll S$, the asymptotic communication cost of PAI is equivalent to traditional methods.

\subsubsection{Active refinement costs}
Active learning involves GP training and optimization of the acquisition function, but only a small number of likelihood evaluations. Thus, under the embarrassingly parallel MCMC assumption that likelihood evaluations are costly (e.g., due to large datasets), active learning is relatively inexpensive \citep{acerbi2018variational}. More importantly, as shown in our ablation study in Appendix \ref{sec:ablation}, this step is crucial to avoid the pathologies of embarrassingly parallel MCMC.

\subsubsection{Sampling complexity}
Sampling from the aggregate approximate posterior $q(\theta)$ only requires evaluating the GP predictive mean for each subposterior and does not require access to the data or all samples. The sampling cost is linear in the number of subposteriors $K$ and the size of each GP $\mathcal{O}(D)$. Even if $K$ is chosen to scale as the size of the actual data, each GP only requires a small training set, making them comparably inexpensive.



\begin{figure*}[t!]
\centering
\includegraphics%
[height=8cm]%
{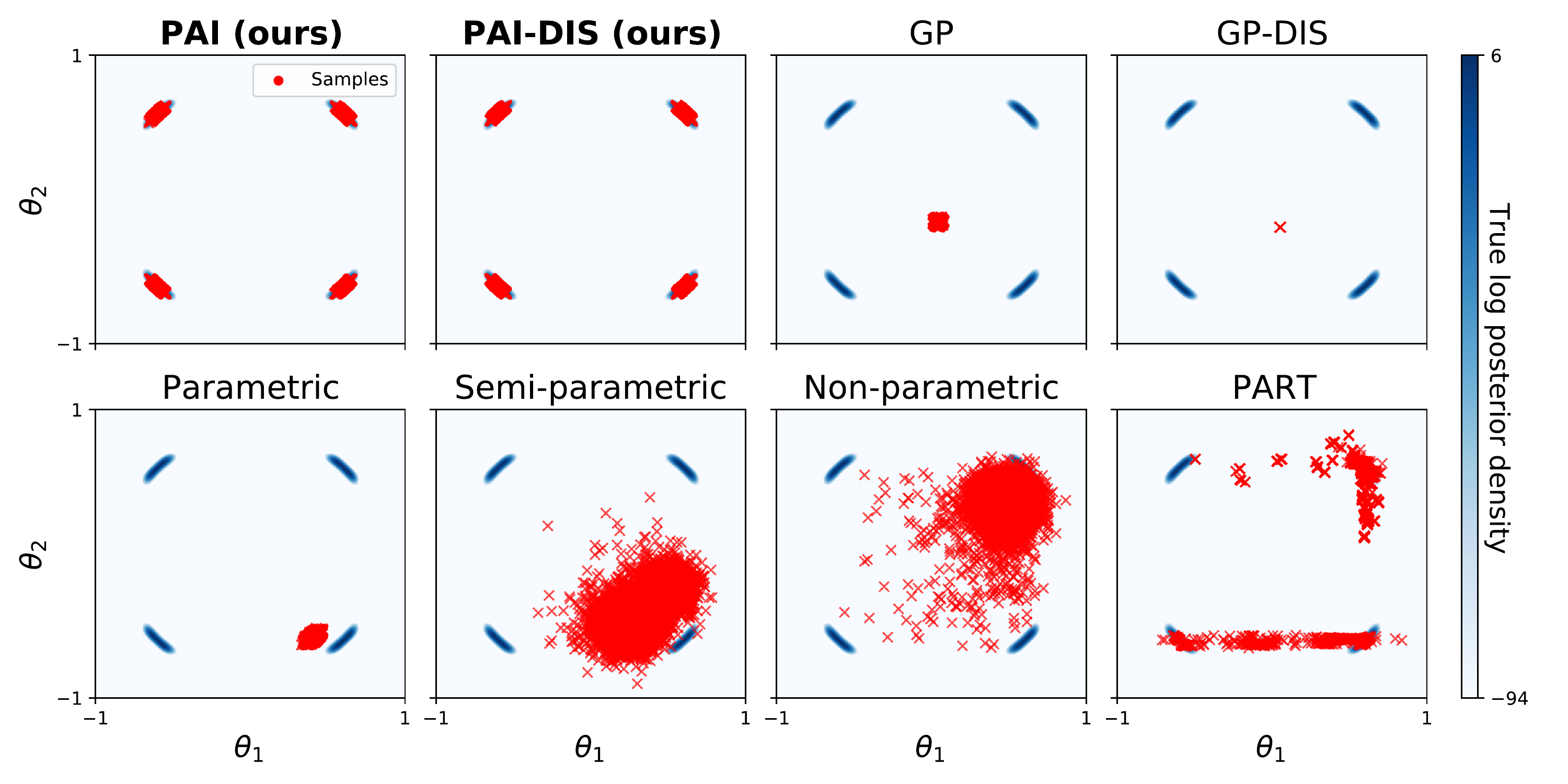}
\vspace{-1em}
  \caption{\textbf{Multi-modal posterior.} Each panel shows samples from the combined approximate posterior (red) against the ground truth (blue). With exception of PAI, all methods completely miss at least one high-density region. Moreover, PAI is the only method that does not assign mass to regions without modes.}
  \label{fig:exp_4modes}
\end{figure*}

\section{EXPERIMENTS}
\label{sec:experiments}

We evaluate PAI on a series of target posteriors with different challenging features. Subsection \ref{subsec:multi_modal} shows results for a posterior with four distinct modes, which is prone to mode collapse (Fig \ref{fig:toy}A).
Subsection \ref{subsec:heavy_tails} targets a posterior with heavy tails, which can lead to underrepresented tails (Fig \ref{fig:toy}C).
Subsection \ref{subsec:rare_events} uses a rare event model to gauge how well our method performs when the true subposteriors are drastically different.
Finally, Subsection \ref{subsec:neuro} concludes with a real-world application to a model and data from computational neuroscience~\citep{acerbi2018bayesian}.
We provide implementation details in Appendix \ref{appendix:implementation} and source code is available at \url{https://github.com/spectraldani/pai}.

\paragraph{Algorithms.} We compare basic PAI and PAI with the optional distributed importance sampling step (PAI-DIS) against six popular and state-of-the-art (SOTA) embarrassingly parallel MCMC methods: the parametric, non-parametric and semi-parametric methods by \citet{Neiswanger2014}; PART \citep{Wang2015}; and two other GP-surrogate methods \citep{Nemeth2018}, one using a simple combination of GPs (GP) and the other using the distributed importance sampler (GP-DIS; see Section \ref{sec:past}).

\paragraph{Procedure.}
For each problem, we randomly split the data in equal-sized partitions and divide the target posterior into $K$ subposteriors (Eq. \ref{eq:pde}). We run MCMC separately on each subposterior using Stan with multiple chains \citep{Carpenter2017}. The same MCMC output is then processed by the different algorithms described above, yielding a combined approximate posterior for each method.
%
To assess the quality of each posterior approximation, we compute the mean marginal total variation distance (MMTV), the 2-Wasserstein (W2) distance, and the Gaussianized symmetrized Kullback-Leibler (\gskl) divergence between the approximate and the true posterior, with each metric focusing on different features.
%
For each problem, we computed ground-truth posteriors using numerical integration (for $D \le 2$) or via extensive MCMC sampling in Stan \citep{Carpenter2017}.
For all GP-based methods (PAI, PAI-DIS, GP, GP-DIS), we sampled from the potentially multimodal combined GP (Eq. \ref{eq:sumgp}) using importance sampling/resampling with an appropriate proposal.
We report results as mean $\pm$ standard deviation across ten runs in which the entire procedure was repeated with different random seeds. For all metrics, lower is better, and the best (statistically significant) results for each metric are reported in bold. See Appendix \ref{supp:evaluation} for more details.

\subsection{Multi-modal posterior}
\label{subsec:multi_modal}

\paragraph{Setting.}
In this synthetic example, the data consist of $N=10^{3}$ samples $y_1, \ldots, y_N$ drawn from the following hierarchical model:
\begin{equation*}
\label{eq:multimodal}
\begin{split}
    \theta \sim p(\theta) & = \mathcal{N}(0, \sigma_p^2 \mathbb{I}_2)\\
    y_1, \ldots, y_N \sim p(y_n | \theta) & =  \sum_{i=1}^{2}\frac{1}{2}\mathcal{N}\left(P_i(\theta_i), \sigma^2_l\right)
\end{split}
\end{equation*}
where $\theta \in \mathbb{R}^2$, $\sigma_p=\sigma_l=1/4$ and $P_i$'s are second-degree polynomial functions. By construction, the target posterior $p(\theta | y_1, \dots, y_N) \propto p(\theta) \prod_{n=1}^N p(y_n | \theta)$ is multimodal with four modes. We run parallel inference on $K=10$ equal-sized partitions of the data. We provide more details regarding $P_1, P_2$ in Appendix \ref{supp:models}.

\paragraph{Results.} Fig \ref{fig:exp_4modes} shows the output of each parallel MCMC method for a typical run, displayed as samples from the approximate combined posterior overlaid on top of the ground-truth posterior. Due to MCMC occasionally missing modes in subposterior sampling, the combined posteriors from all methods except PAI lack at least one mode of the posterior (\emph{mode collapse}, as seen in Fig \ref{fig:toy}A).
Other methods also often inappropriately distribute mass in low-density regions (as seen in Fig \ref{fig:toy}B). In contrast, PAI accurately recovers all the high-density regions of the posterior achieving a near-perfect match. Table \ref{tab:multi-modal} shows that PAI consistently outperforms the other methods in terms of metrics.
%
%

\begin{table}[h!]
    \caption{\textbf{Multi-modal posterior.} 
    }
    \centering
\resizebox{\columnwidth}{!}{
    \begin{tabular}{l c c c }
    \toprule
    \bf{Model}  &  \bf{MMTV} & \bf{W2} & \bf{\gskl}\\
    \toprule
    Parametric&
        $\meanstd{0.89}{0.12}$&  
        $\meanstd{1.08}{0.33}$&  
        $\meanstd[2]{8.9}{11}$\\ 
    Semi-param.&
        $\meanstd{0.81}{0.09}$&  
        $\meanstd{1.08}{0.12}$&  
        $\meanstd[1]{5.6}{1.3}$\\ 
    Non-param.&
        $\meanstd{0.81}{0.09}$&  
        $\meanstd{1.12}{0.09}$&  
        $\meanstd[1]{5.0}{1.8}$\\ 
    PART&
        $\meanstd{0.55}{0.09}$&  
        $\meanstd{1.06}{0.33}$&  
        $\meanstd[2]{7.3}{14}$\\ 
    GP&
        $\meanstd{0.93}{0.16}$&  
        $\meanstd{1.01}{0.36}$&  
        $\meanstd[4]{1.2}{1.3}$\\ 
    GP-DIS&
        $\meanstd{0.87}{0.18}$&  
        $\meanstd{1.04}{0.34}$&  
        $\meanstd[16]{4.8}{14}$\\ 
    \midrule
    \textbf{PAI}&
        $\mathbf{\meanstd{0.037}{0.011}}$&  
        $\mathbf{\meanstd{0.028}{0.011}}$&  
        $\mathbf{\meanstd[-4]{1.6}{1.7}}$\\ 
    \textbf{PAI-DIS}&
        $\mathbf{\meanstd{0.034}{0.019}}$&  
        $\mathbf{\meanstd{0.026}{0.008}}$&  
        $\mathbf{\meanstd[-5]{3.9}{2.4}}$\\ 
    \bottomrule
    \end{tabular}
}
    \label{tab:multi-modal}
\end{table}

\paragraph{Large datasets.} To illustrate the computational benefits of using PAI for larger datasets, we repeated the same experiment in this section but with $10^5$ data points in each of the $K=10$ partitions. Remarkably, even for this moderate-sized dataset, we notice a $6\times$ speed-up -- decreasing the total running time from 6 hours to 57 minutes, (50 for subposterior sampling + 7 from PAI; see Appendix  \ref{appendix:large_dataset}).
Overall, PAI's running time is in the same order of magnitude as the previous SOTA \citep[e.g.][]{Wang2015, Nemeth2018}.
However, only PAI returns correct results while other methods fail.


\begin{figure}[t!]
\centering
\includegraphics%
[width=0.9\linewidth]%
{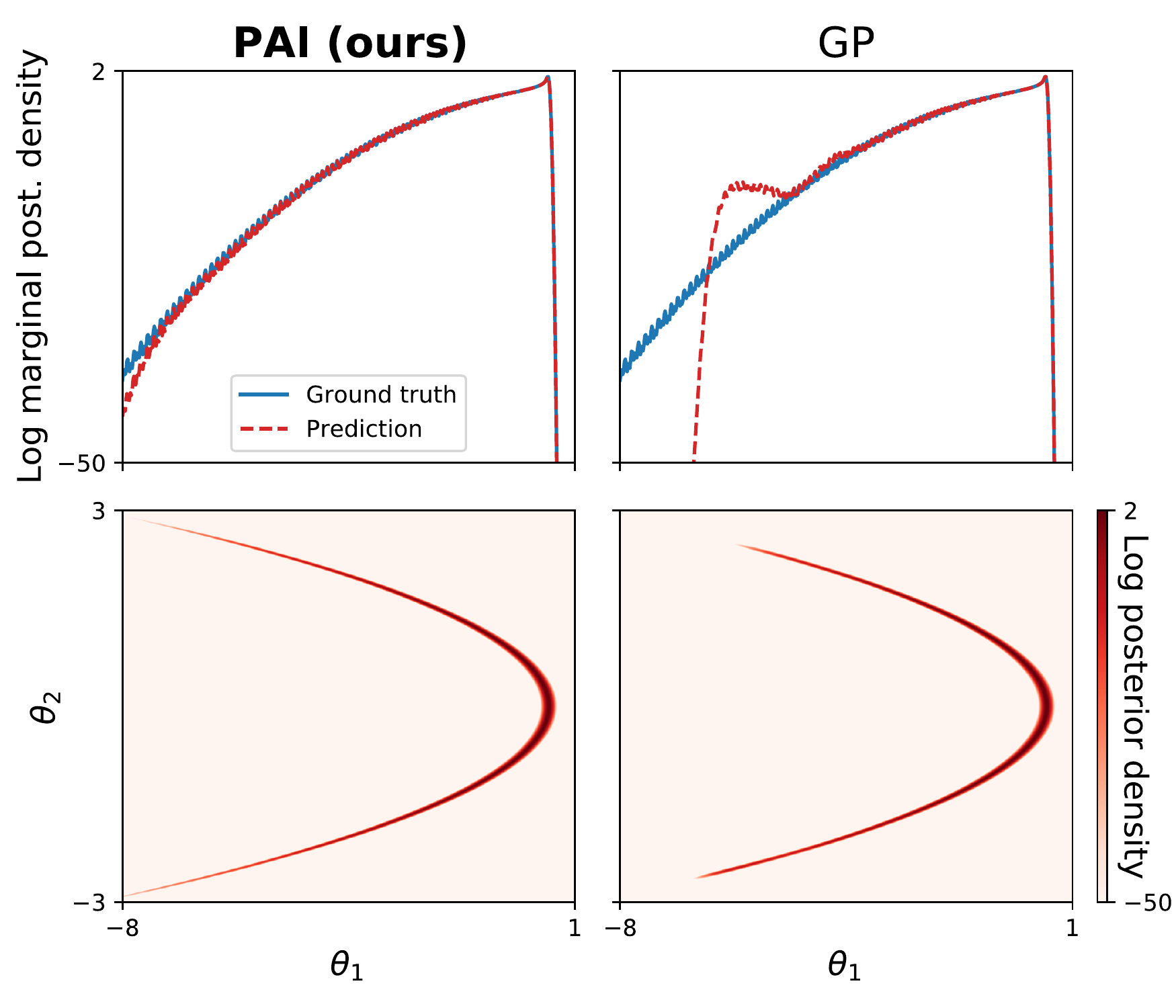}
\vspace{-1em}
    \caption{\textbf{Warped Student's t.} Top: Log marginal posterior for $\theta_1$. Bottom: Log posterior density. Thanks to active sampling, PAI better captures details in the depths of the tails. }
  \label{fig:warped}
\end{figure}

\subsection{Warped Student's t}
\label{subsec:heavy_tails}

\paragraph{Setting.} We now turn to a synthetic example with heavy tails. Consider the following hierarchical model:
\begin{equation*}
\begin{split}
    \theta \sim p(\theta) & =  \mathcal{N}(0, \sigma^2_p \mathbb{I}_2)\\
    y_1, \ldots, y_N \sim p(y_n | \theta) & = \mathrm{StudentT}\left(\nu, \theta_1 + \theta_2^2, \sigma_l^2\right)
\end{split}
\end{equation*}
where $\theta \in \mathbb{R}^2$, $\nu=5$ is the degrees of freedom of the Student's $t$-distribution, $\sigma_p=1$, and $\sigma_l=\sqrt{2}$. This model is a heavy-tailed variant of the Warped Gaussian model studied in earlier work, e.g., \citet{Nemeth2018, Mesquita2019}.
 As before, we generate $N=10^{3}$ samples and split the data into $K=10$ partitions for parallel inference.

\paragraph{Results.} Fig \ref{fig:warped} shows the full posterior and the marginal posterior for $\theta_1$ obtained using the two best-performing methods without DIS refinement, PAI and GP (see Table \ref{tab:warped-gaussian}). While PAI(-DIS) is very similiar to GP(-DIS) in terms of metrics, Fig \ref{fig:warped} shows that, unlike GP(-DIS), PAI accurately captures the far tails of the posterior which could be useful for downstream tasks, avoiding failure mode III (Fig \ref{fig:toy}C).

\begin{table}[h]
    \caption{\textbf{Warped Student's t.}}
    \centering
\resizebox{\columnwidth}{!}{
    \begin{tabular}{l c c c }
    \toprule
    \bf{Model}  &  \bf{MMTV} & \bf{W2} & \bf{\gskl}\\
    \toprule
    Parametric&
        $\meanstd{0.51}{0.01}$&  
        $\meanstd{0.71}{0.07}$&  
        $\meanstd[0]{1.9}{0.1}$\\ 
    Semi-param.&
        $\meanstd{0.50}{0.03}$&  
        $\meanstd{0.57}{0.05}$&  
        $\meanstd[1]{1.1}{0.2}$\\ 
    Non-param.&
        $\meanstd{0.51}{0.02}$&  
        $\meanstd{0.59}{0.03}$&  
        $\meanstd[1]{1.2}{0.2}$\\ 
    PART&
        $\meanstd{0.66}{0.07}$&  
        $\meanstd{0.78}{0.09}$&  
        $\meanstd[2]{1.2}{0.7}$\\ 
    GP&
        $\mathbf{\meanstd{0.015}{0.003}}$&  
        $\mathbf{\meanstd{0.003}{0.002}}$&  
        $\meanstd[-4]{4.5}{10.5}$\\ 
    GP-DIS&
        $\meanstd{0.018}{0.004}$&  
        $\mathbf{\meanstd{0.002}{0.001}}$&  
        $\meanstd[-5]{6.6}{5.8}$\\ 
    \midrule
    \textbf{PAI}&
        $\mathbf{\meanstd{0.015}{0.003}}$&  
        $\mathbf{\meanstd{0.002}{0.001}}$&  
        $\mathbf{\meanstd[-5]{1.2}{0.8}}$\\ 
    {PAI-DIS}&
        ${\meanstd{0.018}{0.003}}$&  
        $\mathbf{\meanstd{0.002}{0.001}}$&  
        $\meanstd[-5]{3.8}{3.4}$\\ 
    \bottomrule
    \end{tabular}
}
    \label{tab:warped-gaussian}
\end{table}

\subsection{Rare categorical events}
\label{subsec:rare_events}

\paragraph{Setting.} To evaluate how our method copes with heterogeneous subposteriors, we run parallel inference for a synthetic example with Categorical likelihood and $N=10^3$ discrete observations split among three classes. To enforce heterogeneity, we make the first two classes rare (true probability $\theta_1 = \theta_2 = 1/N$) and the remaining class much more likely (true probability $\theta_3 = (N-2)/N$). Since we split the data into $K=10$ equal-sized parts, some of them will not contain even a single rare event. We perform inference over $\theta \in \Delta^2$ (probability 2-simplex) with a symmetric Dirichlet prior with concentration parameter $\alpha = 1/3$.

\begin{figure*}[t!]
\centering
\includegraphics%
[height=7cm]%
{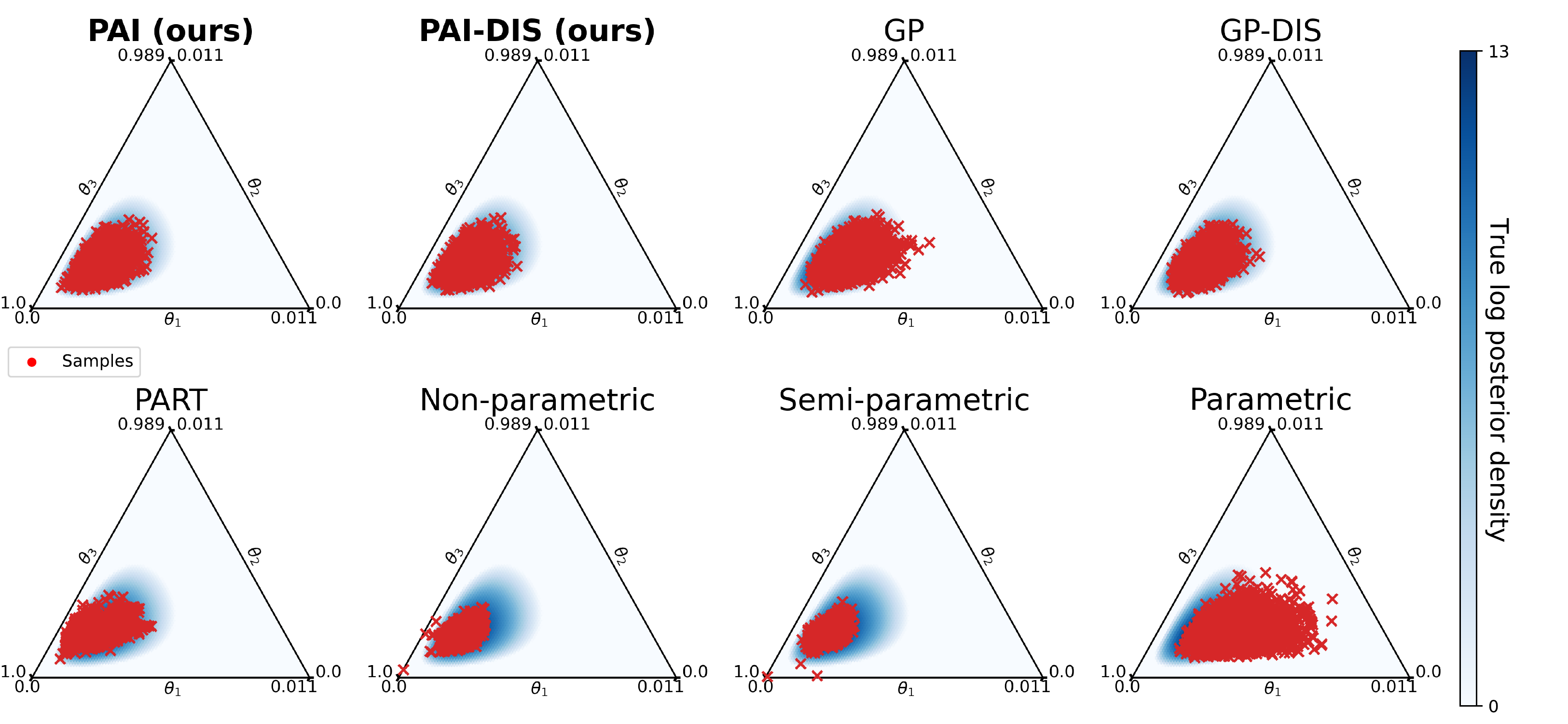}
\vspace{-1em}
  \caption{\textbf{Rare categorical events.} Each ternary plot shows  samples from the combined approximate posterior (red) on top of the true posterior (blue). Note that the panels are zoomed in on the relevant corner of the probability simplex. Of all methods, PAI is the one that best captures the shape of the posterior. }
  \label{fig:rarecat}
\end{figure*}

\paragraph{Results.}  Fig \ref{fig:rarecat} shows the samples from the combined approximate posterior for each method. In this example, PAI-DIS matches the shape of the target posterior extremely well, followed closely by GP-DIS (see also Table \ref{tab:rare}). Notably, even standard PAI (without the DIS correction) produces a very good approximation of the posterior -- a further display of the ability of PAI of capturing fine details of each subposterior, particularly important here in the combination step due to the heterogeneous subposteriors. By contrast, the other methods end up placing excessive mass in very-low-density regions (PART, Parametric, GP) or over-concentrating (Non-parametric, Semi-parametric).
%

\begin{table}[h]
    \caption{\textbf{Rare categorical events.}
    }
    \centering
\resizebox{\columnwidth}{!}{
    \begin{tabular}{l c c c }
    \toprule
    \bf{Model}  &  \bf{MMTV} & \bf{W2} & \bf{\gskl}\\
    \toprule
    Parametric&
        $\meanstd{0.26}{0.14}$&  
        $\meanstd{0.15}{0.19}$&  
        $\meanstd[0]{1.1}{1.4}$\\ 
    Semi-param.&
        $\meanstd{0.49}{0.21}$&  
        $\meanstd{0.27}{0.23}$&  
        $\meanstd[0]{3.5}{3.4}$\\ 
    Non-param.&
        $\meanstd{0.43}{0.17}$&  
        $\meanstd{0.19}{0.25}$&  
        $\meanstd[0]{2.8}{3.9}$\\ 
    PART&
        $\meanstd{0.31}{0.14}$&  
        $\meanstd{0.08}{0.13}$&  
        $\meanstd[-1]{8.6}{10}$\\ 
    GP&
        $\meanstd{0.16}{0.09}$&  
        $\meanstd{0.04}{0.07}$&  
        $\meanstd[-1]{3.5}{4.8}$\\ 
    GP-DIS&
        $\meanstd{0.011}{0.002}$&  
        $\meanstd[-4]{6.3}{0.9}$&  
        $\meanstd[-4]{1.1}{1.5}$\\ 
    \midrule
    PAI&
        ${\meanstd{0.028}{0.027}}$&  
        ${\meanstd{0.001}{0.002}}$&  
        ${\meanstd[-3]{8.0}{16}}$\\ 
    \textbf{PAI-DIS}&
        $\mathbf{\meanstd{0.009}{0.002}}$&  
        $\mathbf{\meanstd[-4]{5.4}{0.8}}$&  
        $\mathbf{\meanstd[-5]{4.3}{2.1}}$\\ 
    \bottomrule
    \end{tabular}
}
    \label{tab:rare}
\end{table}

\subsection{Multisensory causal inference}
\label{subsec:neuro}

\paragraph{Setting.}
\emph{Causal inference} (CI) in multisensory perception denotes the process whereby the brain decides whether distinct sensory cues come from the same source, a commonly studied problem in computational and cognitive neuroscience \citep{kording2007causal}.
%
%
%
%
%
Here we compute the posterior for a 6-parameter CI model given the data of subject S1 from \citep{acerbi2018bayesian} (see Appendix \ref{supp:models} for model details).
The fitted model is a proxy for a large class of similar models that would strongly benefit from parallelization due to likelihoods that do not admit analytical solutions, thus requiring costly numerical integration. 
For this experiment, we run parallel inference over $K = 5$ partitions of the $N = 1069$ observations in the dataset.

\paragraph{Results.} Table \ref{tab:neuroscience} shows the outcome metrics of parallel inference. Similarly to the rare-events example, we find that PAI-DIS obtains an excellent approximation of the true posterior, with GP-DIS performing about equally well (slightly worse on the \gskl metric). Despite lacking the DIS refinement step, standard PAI performs competitively, achieving a reasonably good approximation of the true posterior (see Appendix \ref{supp:models}). All the other methods perform considerably worse; in particular the GP method without the DIS step is among the worst-performing methods on this example.

\begin{table}[h]
    \caption{\textbf{Multisensory causal inference.}}
    \centering
\resizebox{\columnwidth}{!}{
    \begin{tabular}{l c c c }
    \toprule
    \bf{Model}  &  \bf{MMTV} & \bf{W2} & \bf{\gskl}\\
    \toprule
    Parametric&
        $\meanstd{0.40}{0.05}$&  
        $\meanstd{4.8}{0.6}$&  
        $\meanstd[1]{1.2}{0.4}$\\ 
    Semi-param.&
        $\meanstd{0.68}{0.07}$&  
        $\meanstd{9.7}{9.6}$&  
        $\meanstd[1]{5.6}{3.2}$\\ 
    Non-param.&
        $\meanstd{0.26}{0.02}$&  
        $\meanstd{0.52}{0.14}$&  
        $\meanstd[0]{5.3}{0.3}$\\ 
    PART&
        $\meanstd{0.24}{0.04}$&  
        $\meanstd{1.5}{0.5}$&  
        $\meanstd[0]{8.0}{5.4}$\\ 
    GP&
        $\meanstd{0.49}{0.25}$&  
        $\meanstd{17}{23}$&  
        $\meanstd[1]{6.3}{8.9}$\\ 
    GP-DIS&
        $\mathbf{\meanstd{0.07}{0.03}}$&  
        $\mathbf{\meanstd{0.16}{0.07}}$&  
        $\meanstd[-1]{8.7}{14}$\\ 
    \midrule
    {PAI}&
        ${\meanstd{0.16}{0.04}}$&  
        ${\meanstd{0.56}{0.21}}$&  
        ${\meanstd[0]{2.0}{1.7}}$\\ 
    \textbf{PAI-DIS}&
        $\mathbf{\meanstd{0.05}{0.04}}$&  
        $\mathbf{\meanstd{0.14}{0.13}}$&  
        $\mathbf{\meanstd[-1]{2.9}{3.6}}$\\ 
    \bottomrule
    \end{tabular}
} \vspace{-0.5em}
    \label{tab:neuroscience}
\end{table}


\section{RELATED WORKS}
\label{sec:related_works}

While the main staple of embarrassingly parallel MCMC is being a divide-and-conquer algorithm, there are other methods that scale up MCMC using more intensive communication protocols. For instance, \citet{Ahn2014} propose a distributed version of stochatic gradient Langevin dynamics \citep[SGLD,][]{Welling+Teh:2011} that constantly passes around the chain state to computing nodes, making updates only based on local data. However, distributed SGLD tends to diverge from the posterior when the communications are limited, an issue highlighted by recent work \citep{ElMekkaoui2021, Vono2022}. Outside the realm of MCMC, there are also works proposing expectation propagation as a framework for inference on partitioned data \citep{vehtari2020expectation, bui2018partitioned}.

Our method, PAI, builds on top of related work on GP-based surrogate modeling and active learning for log-likelihoods and log-densities. Prior work used GP models and active sampling to learn the intractable marginal likelihood \citep{osborne2012active,gunter2014sampling} or the posterior \citep{kandasamy2015bayesian,wang2017adaptive,jarvenpaa2021parallel}. Recently, the framework of Variational Bayesian Monte Carlo (VBMC) was introduced to simultaneously compute both the posterior and the marginal likelihood \citep{acerbi2018variational,acerbi2019exploration,acerbi2020variational}. PAI extends the above works by dealing with  partitioned data in the embarrassingly parallel setting, similarly to \citet{Nemeth2018}, but with the key addition of active learning and other algorithmic improvements.

\section{DISCUSSION}
\label{sec:discussion}

In this paper, we first exposed several potential major failure modes of existing embarrassingly parallel MCMC methods. We then proposed a solution with our new method, \emph{parallel active inference} (PAI), which incorporates two key strategies: sample sharing and active learning. On a series of challenging benchmarks, we demonstrated that `vanilla' PAI is competitive with current state-of-the-art parallel MCMC methods and deals successfully with scenarios (e.g., multi-modal posteriors) in which all other methods catastrophically fail. When paired with an optional refinement step (PAI-DIS), the proposed method consistently performs on par with or better than state-of-the-art. Our results show the promise of the proposed strategies to deal with the challenges arising in parallel MCMC. Still, the solution is no silver bullet and several aspects remain open for future research.

\subsection{Limitations and future work}

The major limitation of our method, a common problem to surrogate-based approaches, is scalability to higher dimensions. Most GP-based approaches for Bayesian posterior inference are limited to up to $\sim 10$ dimensions, see e.g. \citealp{acerbi2018variational,acerbi2020variational,jarvenpaa2021parallel}. Future work could investigate methods to scale GP surrogate modeling to higher dimensions, for example taking inspiration from high-dimensional approaches in Bayesian optimization (e.g., \citealp{kandasamy2015high}).

More generally, the validity of any surrogate modeling approach hinges on the ability of the surrogate model to faithfully represent the subposteriors. Active learning helps, but model mismatch in our method is still a potential issue that hints at future work combining PAI with more flexible surrogates such as GPs with more flexible kernels \citep{wilson2013gaussian} or deep neural networks \citep{Mesquita2019}. For the latter, obtaining the uncertainty estimates necessary for active learning would involve Bayesian deep learning techniques (e.g., \citealp{maddox2019simple}).

As discussed before, our approach is not `embarrassingly' parallel in that it requires a mandatory global communication step in the sample sharing part (see Section \ref{sec:samplesharing}). The presence of additional communication steps seem inevitable to avoid catastrophic failures in parallel MCMC, and has been used before in the literature (e.g., the DIS step of \citealp{Nemeth2018}). Our method affords an optional final refinement step (PAI-DIS) which also requires a further global communication step. At the moment, there is no automated diagnostic to determine whether the optional DIS step is needed. Our results show that PAI already performs well without DIS in many cases. Still, future work could include an analysis of the GP surrogate uncertainty to recommend the DIS step when useful.

\section*{Acknowledgments}

This work was supported by the Academy of Finland (Flagship programme: Finnish Center for Artificial Intelligence FCAI and grants 328400, 325572) and UKRI (Turing AI World-Leading Researcher Fellowship, EP/W002973/1). We also acknowledge the computational resources provided by the Aalto Science-IT Project from Computer Science IT.

{
\small
\bibliographystyle{plainnat} 
\bibliography{references}
}


\setcounter{footnote}{0}
\setcounter{figure}{0}
\setcounter{table}{0}
\setcounter{equation}{0}
\setcounter{algorithm}{0}
\renewcommand{\theequation}{S\arabic{equation}}
\renewcommand{\thetable}{S\arabic{table}}
\renewcommand{\thefigure}{S\arabic{figure}}
\renewcommand{\thealgorithm}{S\arabic{algorithm}}

\clearpage
\appendix
\thispagestyle{empty}
\onecolumn \makesupplementtitle

In this Supplement, we include extended explanations, implementation details, and additional results omitted from the main text.

Code for our algorithm and to generate the results in the paper is available at:
\url{https://github.com/spectraldani/pai}.

\section{Failure modes of embarrassingly parallel MCMC explained}
\label{supp:failure_modes}

In Section 2.1 of the main text we presented three major failure modes of embarrassingly parallel MCMC (Markov Chain Monte Carlo) methods. These failure modes are illustrated in Fig \ref{fig:toy} in the main text. In this section, we further explain these failure types going through Fig \ref{fig:toy} in detail.

\subsection{Mode collapse (Fig 1A)}
\label{sec:failure_1}
Fig \ref{fig:toy}A illustrates \emph{mode collapse}. In this example, the true subposteriors $p_1$ and $p_2$ both have two modes (see Fig \ref{fig:toy}A, top two panels). However, while in $p_1$ the two modes are relatively close to each other, in $p_2$ they are farther apart. Thus, when we run an MCMC chain on $p_2$, it gets stuck into a single high-density region and is unable to jump to the other mode (`unexplored mode', shaded area).
This poor exploration of $p_2$ is fatal to any standard combination strategy used in parallel MCMC, erasing entire regions of the posterior (Fig \ref{fig:toy}A, bottom panel). While in this example we use PART \citep{Wang2015}, Section \ref{subsec:multi_modal} shows this common pathology in several other methods as well. Our proposed solution to this failure type is sample sharing (see Section \ref{sec:step2}).

\subsection{Model mismatch (Fig 1B)}
\label{sec:failure_2}

Fig \ref{fig:toy}B draws attention to \emph{model mismatch} in subposterior surrogates. In this example, MCMC runs smoothly on both subposteriors. However, when we fit regression-based surrogates -- here, Gaussian processes (GPs) -- to the MCMC samples, the behavior of these models away from subposterior samples can be unpredictable. In our example, the surrogate $q_2$ hallucinates a mode that does not exist in the true subposterior $p_2$ (`model hallucination', shaded area in Fig \ref{fig:toy}B). Our proposed solution to correct this potential issue is to explore uncertain areas of the surrogate using active subposterior sampling (see Section \ref{sec:step3}).

In this example, we used a simple GP surrogate approach for illustration purposes. The more sophisticated GP-based Distributed Importance Sampler (GP-DIS; \citet{Nemeth2018}) might seem to provide an alternative solution to model hallucination, in that the hallucinated regions with low true density would be down-weighted by importance sampling. However, the DIS step only works if there are `good' samples that cover regions with high true density that can be up-weighted. If no such samples are present, importance sampling will not work.
As an example of this failure, Fig \ref{fig:exp_4modes} in the main text shows that GP-DIS \citep{Nemeth2018} does not recover from the model hallucination (if anything, the importance sampling step concentrates the hallucinated posterior even more).

\subsection{Underrepresented tails (Fig 1C)}
\label{sec:failure_3}

Fig \ref{fig:toy}C shows how neglecting low-density regions (\emph{underrepresented tails}) can affect the performance of parallel MCMC. In the example, the true subposterior $p_2$ has long tails that are not thoroughly represented by MCMC samples. This under-representation is due both to actual difficulty of the MCMC sampler in exploring the tails, and to mere Monte Carlo noise as there is little (although non-zero) mass in the tails, so the number of samples is low.
Consequently, the surrogate $q_2$ is likely to underestimate the density in this unexplored region, in which the other subposterior $p_1$ has considerable mass.  In this case, even though $q_1$ is a perfect fit to $p_1$, the product $q_1(\theta) q_2(\theta)$ mistakenly attributes near-zero density to the combined posterior in said region (Fig \ref{fig:toy}C, bottom panel). In our method, we address this issue via multiple solutions, in that both sample sharing and active sampling would help uncover underrepresentation of the tails in relevant regions.

For further illustration of the effectiveness of our proposed solutions to these failure modes, see Section \ref{sec:ablation}.

\section{Gaussian processes}
\label{supp:gps}

Gaussian processes (GPs) are a flexible class of statistical models for specifying prior distributions over unknown functions $f : \X \subseteq \mathbb{R}^{\nparams} \rightarrow \mathbb{R}$ \citep{rasmussen2006gaussian}.
In this section, we describe the GP model used in the paper and details of the training procedure.

\subsection{Gaussian process model}
\label{sec:gp_model}

In the paper, we use GPs as surrogate models for log-posteriors (and log-subposteriors), for which we use the following model.
We recall that GPs are defined by a positive definite covariance, or kernel function $\kappa: \X \times \X \rightarrow \mathbb{R}$; a mean function $m: \X \rightarrow \mathbb{R}$; and a likelihood or observation model.

\paragraph{Kernel function.}
For simplicity, we use the common and equivalently-named squared exponential, Gaussian, or exponentiated quadratic kernel,
\begin{equation} \label{eq:cov}
\kappa(\x,\x^\prime; \sigma_f^2, \ell_1, \ldots, \ell_\nparams) = \sigma_f^2 \exp\left[-\frac{1}{2}\left(\x - \x^\prime\right) \bm{\Sigma}_\ell^{-1} \left(\x - \x^\prime\right)^\top\right]  \qquad \text{with} \; \bm{\Sigma}_\ell = \text{diag}\left[\ell_1^2, \ldots, \ell_\nparams^2\right], 
\end{equation}
where $\sigma_f$ is the output length scale and $(\ell_1, \ldots, \ell_\nparams)$ is the vector of input length scales.
Our algorithm does not hinge on choosing this specific kernel and more appropriate kernels might be used depending on the application (e.g., the spectral mixture kernel might provide more flexibility; see \citealp{wilson2013gaussian}).

\paragraph{Mean function.}
When using GPs as surrogate models for log-posterior distributions, it is common to assume a \emph{negative quadratic} mean function such that the surrogate posterior (i.e., the exponentiated surrogate log-posterior) is integrable \citep{Nemeth2018,acerbi2018variational,acerbi2019exploration,acerbi2020variational}. A negative quadratic can be interpreted as a prior assumption that the target posterior is a multivariate normal; but note that the GP can model deviations from this assumption and represent multimodal distributions as well (see for example Fig \ref{fig:exp_4modes} in the main text). In this paper, we use
\begin{equation}
m(\x; m_0, \mu_1, \ldots, \mu_\nparams, \omega_1, \ldots, \omega_\nparams) \equiv m_0 - \frac{1}{2} \sum_{i=1}^{\nparams} \frac{\left(\x_i - \mu_i\right)^2}{\omega_i^2},
\end{equation}
where $m_0$ denotes the maximum, $\left(\mu_1, \ldots, \mu_\nparams\right)$ is the location vector, and $\left(\omega_1, \ldots, \omega_\nparams\right)$ is a vector of length scales.

\paragraph{Observation model.}
Finally, GPs are also characterized by a likelihood or observation noise model. Throughout the paper we assume exact observations of the target log-posterior (or log-subposterior) so we use a Gaussian likelihood with a small variance $\sigmalik^2 = 10^{-3}$ for numerical stability.

\subsection{Gaussian process inference and training}

\paragraph{Inference.}

Conditioned on training inputs $\xx = \left\{\x_1, \ldots,\x_N \right\}$, observed function values $\y = f(\xx)$ and GP hyperparameters $\vgp$, the posterior GP mean and covariance are available in closed form \citep{rasmussen2006gaussian},
\begin{equation} \label{eq:postgp}
\begin{split}
\overline{f}_{\xx,\y}(\x) \equiv \mathbb{E}\left[f(\x) | \xx, \y, \vgp \right] = & \kappa(\x,\xx) \left[\kappa(\xx,\xx) + \sigmalik^2 \mathbb{I}_\nparams \right]^{-1} (\y - m(\xx)) + m(\x) \\
\covar_{\xx,\y}(\x, \x^\prime) \equiv \text{Cov}\left[f(\x), f(\x^\prime) | \xx, \y, \vgp\right] = & \; \kappa(\x,\x^\prime) - \kappa(\x,\xx) \left[\kappa(\xx,\xx) + + \sigmalik^2 \mathbb{I}_\nparams\right]^{-1} \kappa(\xx,\x^\prime),
\end{split}
\end{equation}
where $\vgp$ is a hyperparameter vector for the GP mean, covariance, and likelihood (see Section \ref{sec:gp_model} above); and $\mathbb{I}_\nparams$ is the identity matrix in $\nparams$ dimensions.

\paragraph{Training.}

Training a GP means finding the hyperparameter vector(s) that best represent a given dataset of input points and function observations $\left(\xx, \y\right)$.
In this paper, we train all GP models by maximizing the log marginal likelihood of the GP plus a log-prior term that acts as regularizer, a procedure known as \emph{maximum-a-posteriori} estimation. Thus, the training objective to maximize is
\begin{equation}
\begin{split}
    \log p(\psi | \xx, \y) = & -\frac{1}{2}\left(\y - m(\xx; \vgp)\right)^\top \left[\kappa(\xx, \xx; \vgp) + \sigma^2 \mathbb{I}_\nparams \right]^{-1} \left(\y - m(\xx; \vgp) \right) \\
    & + \frac{1}{2} \log \det \left(\kappa(\xx, \xx; \vgp) + \sigma^2 \mathbb{I}_\nparams  \right) + \log p(\vgp) + \text{const},
\end{split}
\end{equation}
where $p(\vgp)$ is the prior over GP hyperparameters, described below.

\paragraph{Priors.}

We report the prior $p(\vgp)$ over GP hyperparameters in Table \ref{tab:priors}, assuming independent priors over each hyperparameter and dimension $1 \le i \le \nparams$. We set the priors based on broad characteristics of the training set, an approach known as empirical Bayes which can be seen as an approximation to a hierarchical Bayesian model.
\begin{table}
\centering
\begin{tabular}{lll}
Hyperparameter & Description & Prior distribution \\ \hline \\[-1.5ex]
$\log \sigma^2_f$ & Output scale & ---
\\[2ex]
$\log \ell^{(i)}$ & Input length scale & $\mathrm{Log}\mathcal{N}\left(\log\sqrt{\frac{D}{6}}L^{(i)}, \log\sqrt{10^3}\right)$ \\[2ex]
$m_0$ & Mean function maximum & $\mathrm{SmoothBox}\left(y_\text{min}, y_\text{max}, 1.0\right)$ \\[2ex]
$x_m^{(i)}$ & Mean function location & $\mathrm{SmoothBox}\left(B_\text{min}^{(i)}, B_\text{max}^{(i)}, 0.01\right)$ \\[2ex]
$\log \omega^{(i)}$ & Mean function scale & $\mathrm{Log}\mathcal{N}\left(\log\sqrt{\frac{D}{6}}L^{(i)}, \log\sqrt{10^3}\right)$ \\[2ex] \hline
\end{tabular}
\caption{Priors over GP hyperparameters. See text for more details}
\label{tab:priors}
\end{table}
In the table, $\mathbf{B}$ is the `bounding box' defined as the $\nparams$-dimensional box including all the samples observed by the GP so far plus a 10\% margin;
$\mathbf{L} = \mathbf{B}_\text{max}-\mathbf{B}_\text{min}$ is the vector of lengths of the bounding box; and $y_\text{max}$ and $y_\text{min}$ are, respectively, the largest and smallest observed function values of the GP training set. $\mathrm{Log}\mathcal{N}\left(\mu, \sigma\right)$ denotes the log-normal distribution and $\mathrm{SmoothBox}\left(a,b,\sigma\right)$ is defined as a uniform distribution on the interval $[a,b]$ with a Gaussian tail with standard deviation $\sigma$ outside the interval. If a distribution is not specified, we assumed a flat prior.

\subsection{Implementation details}
\label{appendix:implementation}
All GP models in the paper are implemented using GPyTorch\footnote{\url{https://gpytorch.ai/}} \citep{gardner2018gpytorch}, a modern package for GP modeling based on the PyTorch machine learning framework \citep{paszke2019pytorch}. For maximal accuracy, we performed GP computations enforcing exact inference via Cholesky decomposition, as opposed to the asymptotically faster conjugate gradient implementation which however might not reliably converge to the exact solution \citep{maddox2021when}. For parts related to active sampling, we used the BoTorch\footnote{\url{https://botorch.org/}} package for active learning with GPs \citep{Botorch}, implementing the acquisition functions used in the paper as needed.

We used the same GP model and training procedure described in this section for all GP-based methods in the paper, which include our proposed approach and others (e.g., \citealp{Nemeth2018}).

\section{Algorithm details}
\label{supp:algorithm}

\begin{algorithm*}[ht!]
\small
\caption{Parallel Active Inference (PAI)}\label{alg:ours}
\begin{algorithmic}[1]
\INPUT Data partitions $\data_1, \ldots, \data_K$; prior $p(\theta)$; likelihood function $p(\data|\theta)$.
\ParFor{$1 \ldots K$} \Comment{Parallel steps}
\State $\samples_k \leftarrow $ MCMC samples from $p_k(\theta) \propto p(\theta)^{1/K} p(\data_k | \theta)$
\State $\subsamples_k \leftarrow $ \textsc{ActiveSubsample}($\samples_k$) \Comment{See Sections \ref{sec:gpsubposteriors} and \ref{sec:step1}}
\State send $\subsamples_k$ to all other nodes, receive $\subsamples_{\setminus k} = \bigcup_{j\neq k} \subsamples_j$ \Comment{Communication step}
\State $\subsamplestwo_k \leftarrow$ $\subsamples_k \, \cup$ \textsc{SelectSharedSamples}$(\subsamples_k, \subsamples_{\setminus k})$
\Comment{See Sections \ref{sec:samplesharing} and \ref{sec:step2}}
\State $\subsamplesthree_k \leftarrow$ $\subsamplestwo_k \, \cup$ \textsc{ActiveRefinement}$(\subsamplestwo_k)$
\Comment{See Sections \ref{sec:activerefinement} and \ref{sec:step3}}
\State train GP model $\lq_k$ of the log subposterior on $\left(\subsamplesthree_k, \log p_k (\subsamplesthree_k)\right)$
\EndParFor
\State combine subposteriors: $\log q(\theta) = \sum_{k=1}^K \lq_k(\theta) $ \Comment{Centralized step, see Section \ref{sec:combining} 
}
\State \textbf{Optional:} refine $\log q(\theta)$ with Distributed Importance Sampling (DIS) \Comment{See Section \ref{sec:past}}
\end{algorithmic}
\end{algorithm*}
In this section, we describe additional implementation details for the steps of the Parallel Active Inference (PAI) algorithm introduced in the main text. Algorithm \ref{alg:ours} illustrates the various steps. Each function called by the algorithm may involve several sub-steps (e.g., fitting interim surrogate GP models) which are detailed in the following sections.

\subsection{Subposterior modeling via GP regression}
\label{sec:step1}

Here we expand on Section \ref{sec:gpsubposteriors} in the main text. We recall that at this step each node $k \in \{1, \ldots, K \}$ has run MCMC on the local subposterior $p_k$, obtaining a set of samples $\samples_k$ and their log-subposterior values $\log p_k\left(\samples\right)$.

The goal now is to `thin' the samples to a subset which is still very informative of the shape of the subposterior, so that it can be used as a training set for the GP surrogate. The rationale is that using all the samples for GP training is expensive in terms of both computational and communication costs, and it can lead to numerical instabilities. In previous work, \citet{Nemeth2018} have used \emph{random} thinning which is not guaranteed to keep relevant parts of the posterior.\footnote{Note that instead of thinning we could use sparse GP approximations (e.g., \citealp{titsias2009variational}). However, the interaction between sparse GP approximations and active learning is not well-understood. Moreover, we prefer not to introduce an additional layer of approximation that could reduce the accuracy of our subposterior surrogate models.}

The main details that we cover here are: (1) how we pick the initial subset of samples $\samples^{(0)}_k \subseteq \samples_k$ used to train the initial GP surrogate; (2) how we subsequently perform \emph{active subsampling} to expand the initial subset to include relevant points from $\samples_k$.

\paragraph{Initial subset:} To bootstrap the GP surrogate model, we use a distance clustering method to select the initial subset of samples that we use to fit a first GP model. As this initial subset will be further refined, the main characteristic for selecting the samples is for them to be spread out. For our experiments, we choose a simple $k$-medoids method \citep{park2009kmedoids} with $n_\text{med} = 20\cdot(\nparams+2)$ medoids implemented in the \texttt{scikit-learn-extras} library\footnote{\url{https://github.com/scikit-learn-contrib/scikit-learn-extra}}. The output $n_\text{med}$ medoids represent $\samples^{(0)}_k$.

\paragraph{Active subsampling:} After having selected the initial subset $\samples^{(0)}_k$, we perform $T$ iterations of active subsampling. In each iteration $t + 1$, we greedily select a batch of $n_\text{batch}$ points from the set $\samples_k \setminus \samples_k^{(t)}$ obtained by maximizing a batch version of the maximum interquantile range (MAXIQR) acquisition function, as described by \citet{jarvenpaa2021parallel}; see also main text. The MAXIQR acquisition function has a parameter $u$ which controls the tradeoff between exploitation of regions of high posterior density and exploration of regions with high posterior uncertainty in the GP surrogate. To strongly promote exploration, after a preliminary analysis on a few toy problems, we set $u = 20$ throughout the experiments presented in the paper. In the paper, we set $n_\text{batch} = \nparams$ and the number of iterations $T = 25$, based on a rule of thumb for the total budget of samples required by similar algorithms to achieve good performance at a given dimension (e.g., \citealt{acerbi2018variational,jarvenpaa2021parallel}). The GP model is retrained after the acquisition of each batch.
The procedure locally returns a subset of samples $\subsamples_k = \samples_k^{(T)}$.

\subsection{Sample sharing}
\label{sec:step2}

In this section, we expand on Section \ref{sec:samplesharing} of the main text. We recall that at this step node $k$ receives samples $\subsamples_{\setminus k} = \bigcup_{j \neq k} \subsamples_j$ from the other subposteriors.
To avoid incorporating too many data points into the local surrogate model (for the reasons explained previously), we consider adding a data point to the current surrogate only if: (a) the local model cannot properly predict this additional point; and (b) predicting the exact value would make a difference. If the number of points that are eligible under these criteria is greater than $n_\text{share}$, the set is further thinned using $k$-medoids.

Concretely, let $\theta^\star \in \subsamples_{\setminus k}$ be the data point under consideration. We evaluate the true subposterior log density at the point, $y^\star = \log p_k(\theta^\star)$, and the surrogate GP posterior latent mean and variance at the point, which are, respectively, $\mu_\star = \overline{f}(\theta_\star)$ and $\sigma_\star^2 = C(\theta^\star,\theta^\star)$. We then consider two criteria:
\begin{enumerate}
\item[a.] First, we compute the density of the true value under the surrogate prediction and check if it is above a certain threshold: $\normpdf{\log y^\star}{\mu_\star}{\sigma^2_\star} > R$, where $R=0.01$ in this paper. This criterion is roughly equivalent to including a point only if $\left| \mu_\star - y^\star \right| \gtrsim R^\prime \sigma^\star$, for an appropriate choice of $R^\prime$ (ignoring a sublinear term in $\sigma^\star$), implying that a point is considered for addition if the GP prediction differs from the actual value more than a certain number of standard deviations.
\item[b.] Second, if a point meets the first criterion, we check if the value of the point is actually relevant for the surrogate model.
Let $y_\text{max}$ be the maximum subposterior log-density observed at the current node (i.e., approximately, the log-density at the mode). We exclude a point at this stage if both the GP prediction and the true value $y^\star$ are below the threshold $y_\text{max} - 20 \nparams$, meaning that the point has very low density and the GP correctly predicts that it is very low density (although might not predict the exact value).
\end{enumerate}
Each of the above criteria is checked for all points in $\subsamples_{\setminus k}$ in parallel (the second criterion only for all the points that pass the first one). Note that the second criterion is optional, but in our experiments we found that it increases numerical stability of the GP model, removing points with very low (log) density that are difficult for the GP to handle (for example, \cite{acerbi2018variational,jarvenpaa2021parallel} adopt a similar strategy of discarding very low-density points). More robust GP kernels (see Section \ref{sec:gp_model}) might not need this additional step.

If the number of points that pass both criteria for inclusion is larger than $n_\text{share}$, then $k$-medoids is run on the set of points under consideration with $n_\text{share}$ medoids, where $n_\text{share} = 25 \nparams$. The procedure run at node $k$ locally returns a subset of samples $\subsamplestwo_k$.

\subsection{Active subposterior refinement}
\label{sec:step3}

Here we expand on Section \ref{sec:activerefinement} of the main text. We recall that up to this point, the local GP model of the log-subposterior at node $k$ was trained only using selected subset of samples from the original MCMC runs (local and from other nodes), denoted by $\subsamplestwo_k$.

In this step, each node $k$ locally acquires new points by iteratively optimizing the MAXIQR acquisition function (Eq. \ref{eq:maxiqr} in the main text) over a domain $\mathcal{X} \subseteq \mathbb{R}^\nparams$. For the first iteration, the space $\mathcal{X}$ is defined as the bounding box of $\bigcup_{k} \subsamples_k$ plus a 10\% margin. In other words, $\mathcal{X}$ is initially the hypercube that contains all samples from all subposteriors obtained in the sample sharing step (Section \ref{sec:step2}) extended by a 10\% margin. The limits of $\mathcal{X}$ at the first iteration are computed during the previous stage, without any additional communication cost. In each subsequent iteration, $\mathcal{X}$ is iteratively extended to include the newly sampled points plus a 10\% margin, thus expanding the bounding box if a recently acquired point falls near the boundary.

New points are selected greedily in batches of size  $n_\text{batch} = D$, using the same batch formulation of the MAXIQR acquisition function used in Section \ref{sec:step1}. The local GP surrogate is retrained at the end of each iteration, to ensure that the next batch of points targets the regions of the log-subposterior which are most important to further refine the surrogate.
For the purpose of this work, we repeat the process for $T_\text{active} = 25$ iterations, selected based on similar active learning algorithms \citep{acerbi2018bayesian, jarvenpaa2021parallel}. Future work should investigate an appropriate termination rule to dynamically vary the number of iterations.

The outcome of this step is a final local set of samples $\subsamplesthree_k$ and the log-subposterior GP surrogate model $\mathcal{L}_k$ trained on these samples. Both of these are sent back to the central node for the final combination step.


\section{Experiment details and additional results}

In this section, we report additional results and experimental details omitted from the main text for reasons of space.

\subsection{Ablation study}
\label{sec:ablation}

\begin{figure}
\centering
\includegraphics%
[width=1.0\linewidth]%
{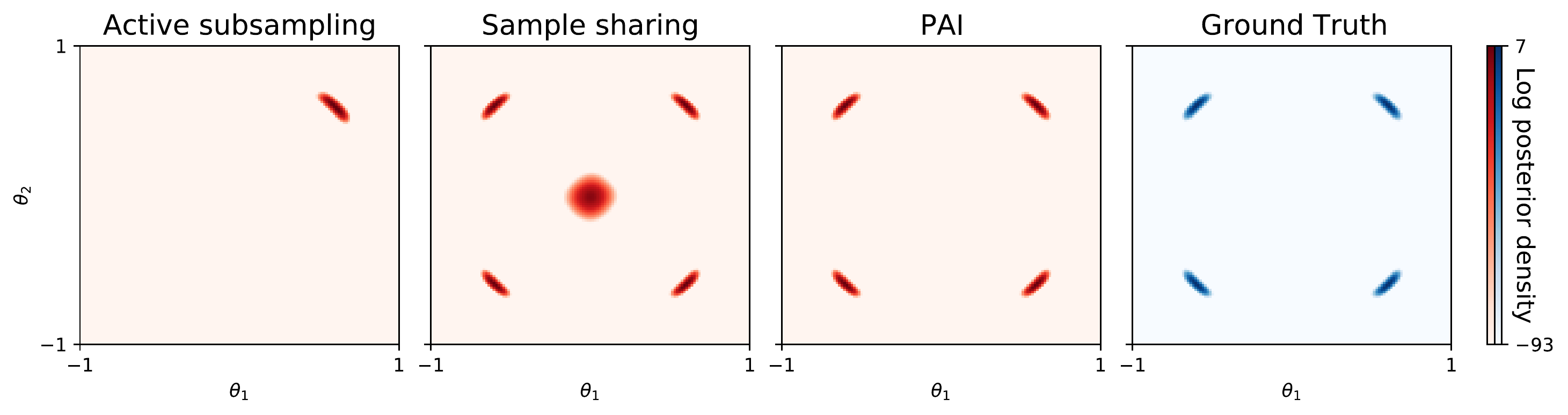}
\vspace{-1em}
  \caption{\textbf{Ablation study for PAI on the multi-modal posterior.} From left to right: The first panel shows the approximate combined posterior density for our method \emph{without} sample sharing and active learning; the second panel uses an additional step to share samples (but no active refinement); the third shows results for full-fledged PAI. The rightmost plot is the ground truth posterior. Note that both sample sharing and active refinement are important steps for PAI: sample sharing helps account for missing posterior regions while active sampling corrects model hallucinations (see text for details).}
\label{fig:ablation}
\end{figure}

As an ablation study, Fig \ref{fig:ablation} breaks down the effect of each step of PAI in the multi-modal posterior experiment from Section \ref{subsec:multi_modal} of the main text.
The first panel shows the full approximate posterior if we were combining it right after active subsampling (Section \ref{sec:step1}), using neither sample sharing nor active refinement.
Note that this result suffers from Failure mode I (mode collapse; see Section \ref{sec:failure_1}), as active subsampling only on the local MCMC samples is not sufficient to recover the missing modes.
The second panel incorporates sample sharing, which covers the missing regions but now suffers from Failure mode II (model mismatch; see Section \ref{sec:failure_2}) with an hallucinated mode in a region where the true posterior has low density.
Finally, the third panel shows full-fledged PAI, which further applies active sampling to explore the hallucinated mode and corrects the density around it. The final result of PAI perfectly matches the ground truth (as displayed in the fourth panel).

\subsection{Performance evaluation}
\label{supp:evaluation}

In this section, we describe in detail the metrics used to assess the performance of the methods in the main text, how we compute these metrics, and the related statistical analyses for reporting our results.

\paragraph{Metrics.}

In the main text, we measured the quality of the posterior approximations via the mean marginal total variation distance (MMTV), 2-Wasserstein (W2) distance, and Gaussianized symmetrized Kullback-Leibler divergence (\gskl) between true and appoximate posteriors. For all metrics, lower is better. We describe the three metrics and their features below:
\begin{itemize}
    \item The MMTV quantifies the (lack of) overlap between true and approximate posterior marginals, defined as
    \begin{equation} \label{eq:mmtv}
        \text{MMTV}(p,q) = \frac{1}{2\nparams} \sum_{d = 1}^\nparams \int_{-\infty}^{\infty} \left|p^\text{M}_d(x_d) - q^\text{M}_d(x_d) \right| dx_d
    \end{equation}
    where $p^\text{M}_d$ and $q^\text{M}_d$ denote the marginal densities of $p$ and $q$ along the $d$-th dimension. Eq. \ref{eq:mmtv} has a direct interpretation in that, for example, a MMTV metric of 0.5 implies that the posterior marginals overlap by 50\% (on average across dimensions). As a rule of thumb, we consider a threshold for a reasonable posterior approximation to be MMTV < 0.2, that is more than $80\%$ overlap.
    \item Wasserstein distances measure the cost of moving amounts of probability mass from one distribution to the other so that they perfectly match -- a commonly-used distance metric across distributions. The W2 metric, also known as \emph{earth mover's} distance, is a special case of Wasserstein distance that uses the Euclidean distance as its cost function. The W2 distance between two density functions $p$ and $q$, with respective supports $\mathcal{X}$ and $\mathcal{Y}$ is given by
    \begin{equation} \label{eq:w2}
        \text{W2}(p, q) = \left[\inf_{T \in \mathcal{T}} \int_{x \in \mathcal{X}} \int_{y \in \mathcal{Y}} \|x - y\|_2 T(x, y)\, dx\, dy\right]^{\frac{1}{2}},
    \end{equation}
    where $\mathcal{T}$ denotes the set of all joint density functions over $\mathcal{X} \times \mathcal{Y}$ with marginals exactly $p$ and $q$. In practice, we use empirical approximations of $p$ and $q$ to compute the W2, which simplifies Eq. \ref{eq:w2} to a linear program.

    \item The \gskl metric is sensitive to differences in means and covariances, being defined as
\begin{equation} \label{eq:gsKL}
\text{GsKL}(p, q) = \frac{1}{2}\left[D_\text{KL}\left(\mathcal{N}[p]||\mathcal{N}[q]\right) + D_\text{KL}(\mathcal{N}[q]|| \mathcal{N}[p])\right],
\end{equation}
    where $D_\text{KL}\left(p||q\right)$ is the Kullback-Leibler divergence between distributions $p$ and $q$ and $\mathcal{N}[p]$ is a multivariate normal distribution with mean equal to the mean of $p$ and covariance matrix equal to the covariance of $p$ (and same for $q$). Eq. \ref{eq:gsKL} can be expressed in closed form in terms of the means and covariance matrices of $p$ and $q$. For reference, two Gaussians with unit variance and whose means differ by $\sqrt{2}$ (resp., $\frac{1}{2}$) have a \gskl of 1 (resp., $\frac{1}{8}$). As a rule of thumb, we consider a desirable target to be (much) less than 1.
\end{itemize}

\paragraph{Computing the metrics.} For each method, we computed the metrics based on samples from the combined approximate posteriors. For methods whose approximate posterior is a surrogate GP model (i.e., GP, GP-DIS, PAI, PAI-DIS in the paper), we drew samples from the surrogate model using importance sampling/resampling \citep{robert2013monte}. As proposal distribution for importance sampling we used a mixture of a uniform distribution over a large hyper-rectangle and a distribution centered on the region of high posterior density (the latter to increase the precision of our estimates). We verified that our results did not depend on the specific choice of proposal distribution.

\paragraph*{Statistical analyses.} For each problem and each method, we reran the entire parallel inference procedure ten times with ten different random seeds.
The same ten seeds were used for all methods -- implying among other things that, for each problem, all methods were tested on the same ten random partitions of the data and using the same MCMC samples on those partitions.
The outcome of each run is a triplet of metrics (MMTV, W2, \gskl) with respect to ground truth. We computed mean and standard deviation of the metrics across the ten runs, which are reported in tables in the main text. For each problem and metric, we highlighted in bold all methods whose mean performance does not differ in a statistically significant way from the best-performing method. Since the metrics are not normally distributed, we tested statistical significance via bootstrap ($n_\text{bootstrap} = 10^6$ bootstrapped datasets) with a threshold for statistical significance of $\alpha = 0.05$.

\subsection{Model details and further plots}
\label{supp:models}

We report here additional details for some of the models used in the experiments in the main paper, and plots for the experiment from computational neuroscience.

\paragraph{Multi-modal posterior.}
In Section \ref{subsec:multi_modal} of the main text we constructed a synthetic multi-modal posterior with four modes. We recall that the generative model is
\begin{equation*}
\label{eq:supp_fourmodes}
\begin{split}
    \theta \sim p(\theta) & = \mathcal{N}(0, \sigma_p^2 \mathbb{I}_2)\\
    y_1, \ldots, y_N \sim p(y_n | \theta) & =  \sum_{i=1}^{2}\frac{1}{2}\mathcal{N}\left(y_n; P_i(\theta_i), \sigma^2_l\right)
\end{split}
\end{equation*}
where $\theta \in \mathbb{R}^2$, $\sigma_p=\sigma_l=1/4$ and $P_i$'s are second-degree polynomial functions.
To induce a posterior with four modes, we chose $P_1(\theta_1)$ and $P_2(\theta_2)$ to be polynomials with exactly two roots, such that, when the observations are drawn from the full generative model in Eq. \ref{eq:supp_fourmodes}, each root will induce a local maximum of the posterior in the vicinity of the root (after considering the shrinkage effect of the prior).
The polynomials are defined as $P_1(x) = P_2(x) = (0.6-x)(-0.6-x)$, so the posterior modes will be in the vicinity of $\theta^\star \in \left\{(0.6, 0.6), (-0.6, 0.6), (0.6, -0.6), (-0.6, -0.6)\right\}$.

\paragraph{Multisensory causal inference.}

\begin{figure}
    \centering
    \includegraphics[width=0.6\linewidth]{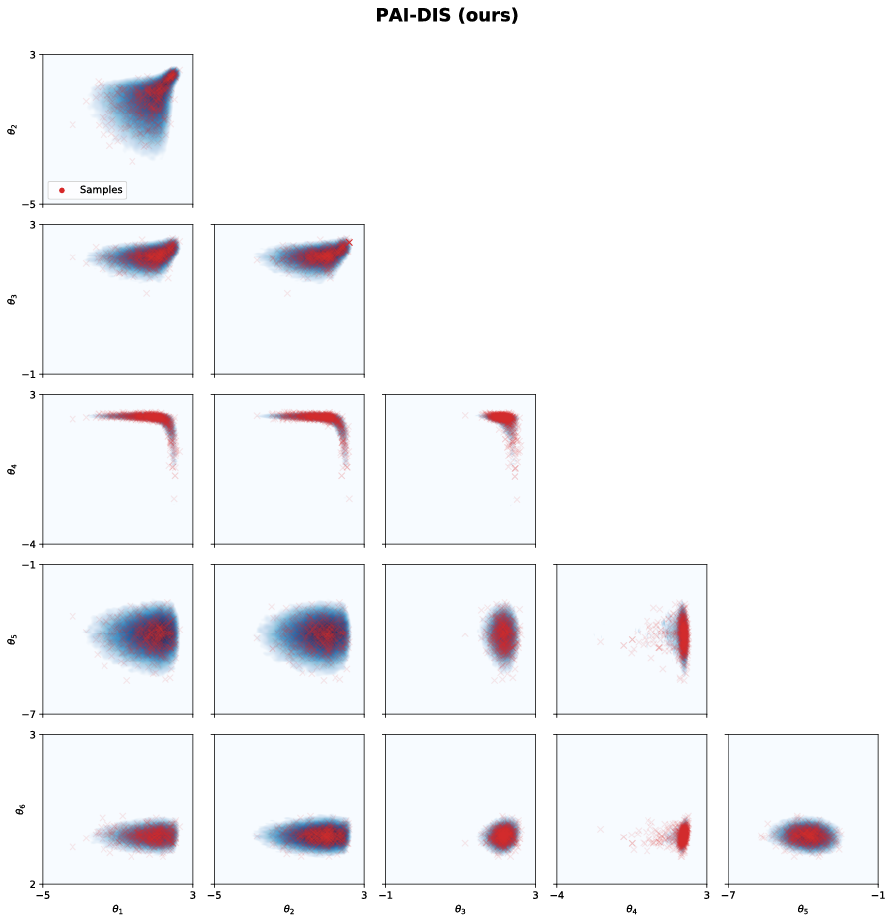}%
    \vspace{12pt}
    \includegraphics[width=0.6\linewidth]{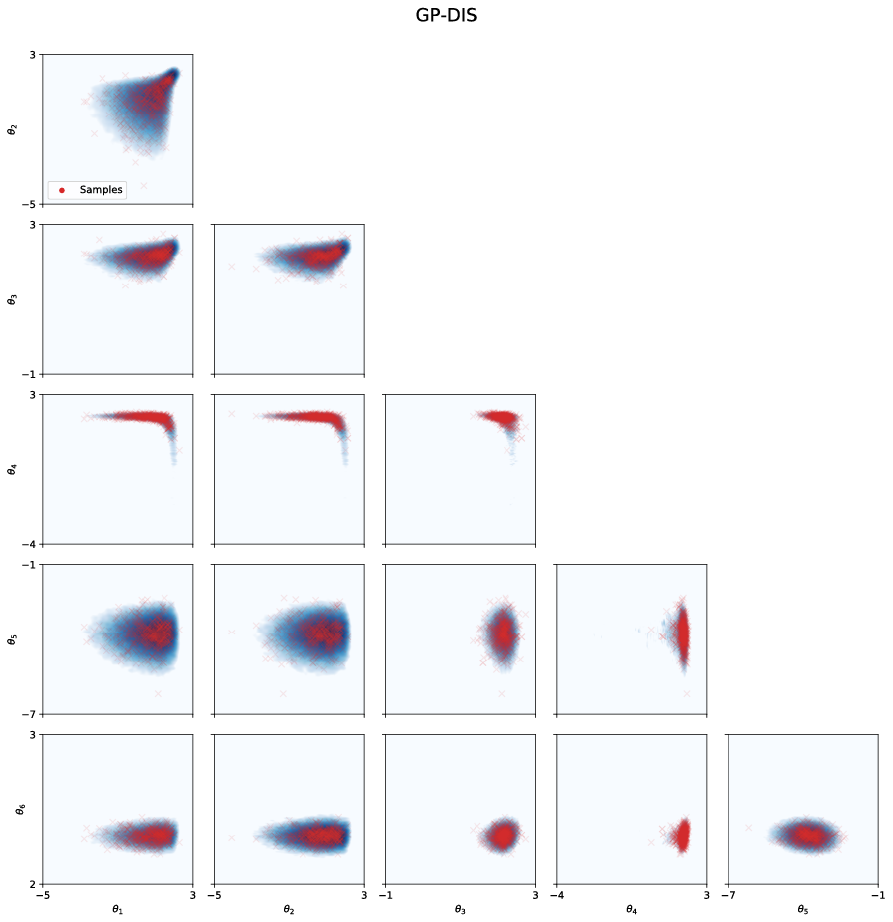}
    \caption{\textbf{PAI-DIS and GP-DIS on the multisensory causal inference task.} Each panel shows two-dimensional posterior marginals as samples from the combined approximate posterior (red) against the ground truth (blue). While PAI-DIS (top figure) and GP-DIS (bottom figure) perform similarly in terms of metrics, PAI-DIS captures some features of the posterior shape more accurately, such as the `boomerang' shape of the $\theta_3$ marginals (middle row).}
    \label{fig:neuro}
\end{figure}

In Section \ref{subsec:neuro} of the main text, we modeled a benchmark visuo-vestibular causal inference experiment \citep{acerbi2018bayesian,acerbi2020variational} which is representative of many similar models and tasks in the fields of computational and cognitive neuroscience.
In the modeled experiment, human subjects, sitting in a moving chair, were asked in each trial whether the direction of movement $s_\text{vest}$ matched the direction $s_\text{vis}$ of a looming visual field. We assume subjects only have access to noisy sensory measurements $z_\text{vest} \sim \mathcal{N}\left(s_\text{vest}, \sigma^2_\text{vest} \right)$, $z_\text{vis} \sim \mathcal{N}\left(s_\text{vis}, \sigma^2_\text{vis}(c) \right)$, where $\sigma_\text{vest}$ is the vestibular noise and $\sigma_\text{vis}(c)$ is the visual noise, with $c \in \{c_\text{low}, c_\text{med}, c_\text{high}\}$  distinct levels of visual coherence adopted in the experiment.
We model subjects' responses with a heuristic `Fixed' rule that judges the source to be the same if $|z_\text{vis} - z_\text{vest}| < \kappa$, plus a probability $\lambda$ of giving a random response \citep{acerbi2018bayesian}. Model parameters are $\theta = (\sigma_\text{vest},\sigma_\text{vis}(c_\text{low}),\sigma_\text{vis}(c_\text{med}),\sigma_\text{vis}(c_\text{high}),\kappa,\lambda)$, nonlinearly mapped to $\mathbb{R}^6$. In the paper, we fit real data from subject S1 of \citep{acerbi2018bayesian}. Example approximate posteriors for the PAI-DIS and GP-DIS methods, the best-performing algorithms in this example, are shown in Fig \ref{fig:neuro}.

\subsection{Scalability of PAI to large datasets}
\label{appendix:large_dataset}

In the main paper, as per common practice in the field, we used moderate dataset sizes ($\sim$10k data points) to easily calculate ground truth.
This choice bears no loss of generality because increasing dataset size only makes subposteriors sharper, which does not increase the difficulty of parallel inference (although more data would not necessarily resolve multimodality, e.g. due to symmetries of the model). On the other hand, small datasets make the reporting of run-times not meaningful, as they are dominated by overheads.

To assess the performance of PAI on large datasets, we ran PAI on the model of Section \ref{subsec:multi_modal}, but now with \emph{1 million data points} ($K = 10$ partitions). Average metrics for PAI were excellent and similar to what we had before: MTV = 0.009, W2 = 0.005, GKL = 2e-05, while all other methods still failed. Moreover, run-times for this experiment illustrate the advantages of using PAI in practice.

We ran experiments using computers equipped with two 8-core Xeon E5 processors and 16GB or RAM each. Here, the total time for parallel inference was 57 minutes --- 50 for subposterior MCMC sampling + 7 for all PAI steps. By contrast, directly running MCMC on the whole dataset took roughly 6 hours.

\end{document}